%% file: main.tex
\documentclass[sigconf,screen,balance]{acmart}

\AtBeginDocument{%
  \providecommand\BibTeX{{%
    \normalfont B\kern-0.5em{\scshape i\kern-0.25em b}\kern-0.8em\TeX}}}

\copyrightyear{2020}
\acmYear{2020}
\setcopyright{acmcopyright}\acmConference[MM '20]{Proceedings of the 28th ACM International Conference on Multimedia}{October 12--16, 2020}{Seattle, WA, USA}
\acmBooktitle{Proceedings of the 28th ACM International Conference on Multimedia (MM '20), October 12--16, 2020, Seattle, WA, USA}
\acmPrice{15.00}
\acmDOI{10.1145/3394171.3413897}
\acmISBN{978-1-4503-7988-5/20/10}

\newcommand{\eat}[1]{}
\usepackage{booktabs}
\usepackage{multirow}
\usepackage{balance}

\usepackage{subfig}
\usepackage{algorithm}

\usepackage{algpseudocode}
\usepackage{amsthm,amsmath,amssymb}
\usepackage{mathrsfs}

\DeclareMathOperator*{\argmin}{arg\,min}

\algnewcommand{\Initialize}[1]{%
  \State \textbf{Initialize:}
  \Statex \hspace*{\algorithmicindent}\parbox[t]{.8\linewidth}{\raggedright #1}
}
\algnewcommand{\Inputs}[1]{%
  \State \textbf{Inputs:}
  \Statex \hspace*{\algorithmicindent}\parbox[t]{.8\linewidth}{\raggedright #1}
}
\algnewcommand{\Outputs}[1]{%
  \State \textbf{Outputs:}
  \Statex \hspace*{\algorithmicindent}\parbox[t]{.8\linewidth}{\raggedright #1}
}

\begin{document}

\title{Adversarial Bipartite Graph Learning for \\Video Domain Adaptation}


\author{Yadan Luo$^1$, Zi Huang$^1$, Zijian Wang$^1$, Zheng Zhang$^{2,3}$, Mahsa Baktashmotlagh$^1$} 
\affiliation{ 
      $^1$The University of Queensland, $^2$Peng Cheng Laboratory, Shenzhen\\
      $^3$Bio-Computing Research Center, Harbin Institute of Technology, Shenzhen
    }
\email{lyadanluol@gmail.com, huang@itee.uq.edu.au, zijian.wang@uq.edu.au, darrenzz219@gmail.com, m.baktashmotlagh@uq.edu.au}

\fancyhead{}
\renewcommand{\shortauthors}{Luo, Y. et al}


%%%%% Version 1
% With the growing popularity of online social media and the scale of registered users, it becomes more important yet challenging to infer the mutual interactions among users. The user interactions are modeled with positive links presenting relations like friendship and negative links indicating relations like antagonism in a signed social network.
\begin{abstract}
Domain adaptation techniques, which focus on adapting models between distributionally different domains, are rarely explored in the video recognition area due to the significant spatial and temporal shifts across the source (i.e. training) and target (i.e. test) domains. As such, recent works on visual domain adaptation which leverage adversarial learning to unify the source and target video representations and strengthen the feature transferability are not highly effective on the videos. To overcome this limitation, in this paper, we learn a domain-agnostic video classifier instead of learning domain-invariant representations, and propose an Adversarial Bipartite Graph (ABG) learning framework which directly models the source-target interactions with a network topology of the bipartite graph. Specifically, the source and target frames are sampled as heterogeneous vertexes while the edges connecting two types of nodes measure the affinity among them. Through message-passing, each vertex aggregates the features from its heterogeneous neighbors, forcing the features coming from the same class to be mixed evenly. Explicitly exposing the video classifier to such cross-domain representations at the training and test stages makes our model less biased to the labeled source data, which in-turn results in achieving a better generalization on the target domain. The proposed framework is agnostic to the choices of frame aggregation, and therefore, four different aggregation functions are investigated for capturing appearance and temporal dynamics. To further enhance the model capacity and testify the robustness of the proposed architecture on difficult transfer tasks, we extend our model to work in a semi-supervised setting using an additional video-level bipartite graph. Extensive experiments conducted on four benchmark datasets evidence the effectiveness of the proposed approach over the state-of-the-art methods on the task of video recognition.
\end{abstract}

\begin{CCSXML}
<ccs2012>
   <concept>
       <concept_id>10010147.10010178.10010224.10010225.10010228</concept_id>
       <concept_desc>Computing methodologies~Activity recognition and understanding</concept_desc>
       <concept_significance>300</concept_significance>
       </concept>
   <concept>
       <concept_id>10010147.10010178</concept_id>
       <concept_desc>Computing methodologies~Transfer learning</concept_desc>
       <concept_significance>500</concept_significance>
       </concept>
 </ccs2012>
\end{CCSXML}
\ccsdesc[500]{Computing methodologies~Transfer learning}
\ccsdesc[300]{Computing methodologies~Activity recognition and understanding}

\keywords{Video Action Recognition; Domain Adaptation.}



\maketitle

\input{1_intro}
\input{2_relatedwork}

\input{3_method}
\input{4_experiments}
\vspace{-0.3cm}
\section{Conclusion}
In this work, we propose a bipartite graph learning framework for unsupervised and semi-supervised video domain adaptation tasks. Different the existing approaches which learn domain-invariant features, we construct a domain-agnostic classifier by leveraging the bipartite graphs to combine the similar source and target features at the training and test time, which helps with reducing the exposure bias. Experiments evidence effectiveness of our proposed approach over the state-of-the-art methods, improving their performance by up to 39.6$\%$ in a semi-supervised setting.

\vspace{-0.2cm}
\begin{acks}
This work was partially supported by ARC DP 190102353.
\end{acks}
\vspace{-0.2cm}

\bibliographystyle{ACM-Reference-Format}
\bibliography{main}

% \appendix

\end{document}

\title{Adversarial Bipartite Graph Learning \\for Video Domain Adaptation}
\eat{\titlenote{Produces the permission block, and
  copyright information}
\subtitle{Supplementary Material}
\subtitlenote{The full version of the author's guide is available as}
  \texttt{acmart.pdf} document}

% Version 1 - Abstract

\maketitle
\fancyhead{}

\section{Algorithm}
The overall algorithm is presented in Algorithm \ref{alg:1}.

\begin{algorithm}[!htb]
	\begin{algorithmic}[1]
		\Inputs{The labeled source data $\mathcal{D}_s = \{(X^s_i, y_i)\}_{i=1}^{N_s}$ and unlabeled target data $\mathcal{D}_t = \{X^t_j\}_{j=1}^{N_t}$;}
		\Outputs{Model parameters ($\theta^*_{fe}, \theta^*_{fv}, \Theta^*_{a}, \theta^*_{ve}, \theta^*_{vn}, \theta^*_l, \theta^*_y$);}
		\Initialize{Hyper-parameters: $\alpha$, $\beta$, $\gamma$, $\lambda$, $K$, $M$;\\
		Minibatch size $B_s, B_t$ and learning rate $\mu$;}
        \For{M epochs}
        \State Update model parameters by descending stochastic gradients according to Equation (13):
        \State $\theta_{d}\gets\theta-\mu\cdot\nabla_{\theta_{d}}(\beta\mathcal{L}_{d});$ $\theta_{l}\gets\theta+\mu\cdot\nabla_{\theta_{l}}(\beta\mathcal{L}_{d});$
        \State $\theta_{y}\gets\theta-\mu\cdot\nabla_{\theta_{y}}(\mathcal{L}_y^s + \gamma\mathcal{L}_y^t - \beta\mathcal{L}_{d});$
        \State $\theta_{vn}\gets\theta-\mu\cdot\nabla_{\theta_{fv}}(\mathcal{L}_y^s + \gamma\mathcal{L}_y^t - \beta\mathcal{L}_{d});$
        \State $\theta_{ve}\gets\theta-\mu\cdot\nabla_{\theta_{ve}}(\mathcal{L} - \beta\mathcal{L}_{d} - \lambda\alpha \mathcal{L}_e^f);$
        \State $\Theta_{a}\gets\theta-\mu\cdot\nabla_{\Theta_{a}}(\mathcal{L} - \beta\mathcal{L}_{d} - \lambda\alpha \mathcal{L}_e^f)$;
        \State $\theta_{fv}\gets\theta-\mu\cdot\nabla_{\theta_{fv}}(\mathcal{L} - \beta\mathcal{L}_{d} - \lambda\alpha \mathcal{L}_e^f)$;
        \State $\theta_{fe}\gets\theta-\mu\cdot\nabla_{\theta_{fe}}(\mathcal{L} - \beta\mathcal{L}_{d})$;
        \State Update the learning rate $\mu$;
        \EndFor
	\end{algorithmic}
	\caption{Pseudo-code of the Proposed ABG Learning.}
	\label{alg:1}
\end{algorithm}

%% file: 1_intro.tex
\section{Introduction}\label{sec:intro}
With the advent of multimedia streaming~\cite{multimedia,multimedia1,multimedia2,multimedia3} and gaming data, automatically recognizing and understanding human actions and events in videos have become increasingly important, especially for practical tasks such as video retrieval~\cite{mm4}, surveillance~\cite{mm3}, and recommendation~\cite{mm,mm1}. Over the past decades, great efforts have been made to boost the recognition performance with deep learning for different purposes including appearances and short-term motions learning ~\cite{DBLP:conf/nips/SimonyanZ14,C3D}, temporal structure modeling~\cite{TSN}, and human skeleton and pose embedding~\cite{skeleton,skeleton1,pose}. While effective, deep learning enables machine recognition at a great cost of labeling large-scale data.
\begin{figure}[!t]
    \centering
    \includegraphics[width=1\linewidth]{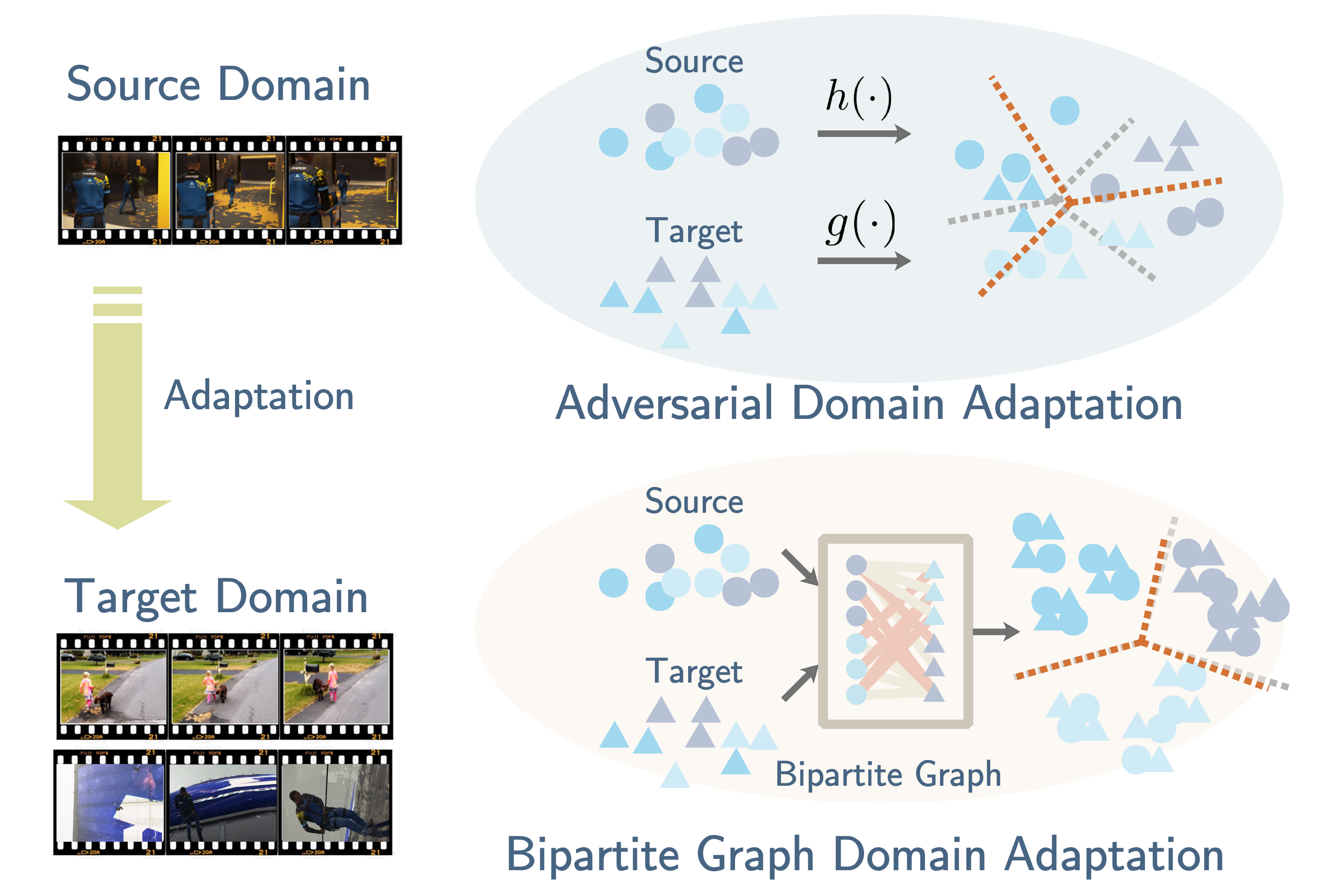}
    \caption{An illustration of the video domain adaptation task. With the unaligned inputs from different domains, the video classifier learned in existing adversarial DA approaches (shown as grey doted lines) can easily overfit the labeled source data. By contrast, the classifier learned using our proposed ABG framework (shown as orange doted lines) is domain-agnostic, which performs equally well at the training and test stages.}
    \label{fig:example}
\end{figure}
To relieve the burden of tedious and expensive labeling, one alternative is to transfer knowledge from the existing annotated training data (i.e. \textbf{source domain}) to the unlabeled or partially labeled test data (i.e. \textbf{target domain}). However, the source and target sets are commonly constructed under varying conditions such as illuminations, camera poses and backgrounds, leading to a huge domain shift. For instance, the Gameplay-Kinetics~\cite{TAN} dataset is built under the challenging ``Synthetic-to-Real'' protocol, where the training videos are synthesized by game engines and the test samples are collected from real scenes. In this case, the domain discrepancy between the source and target domains inevitably leads to a severe degradation of the model generalization performance.

To combat the above dilemmas, domain adaptation (DA) approaches have been investigated to mitigate the domain gap by aligning the distributions across the domains~\cite{MMD,MMD1,MMD_mahsa} or learning domain-invariant representations~\cite{DANN,conditional1}. While the notion of domain adaptation has been widely exploited in the past, the resulting techniques are mostly designed to cope with still images rather than the videos. These image-level DA methods could hardly achieve a good performance on the video recognition tasks as they don't take into account the temporal dependency of the frames when minimizing the discrepancy between the domains.

% Summarize Video DA and highlight the main challenges
% (1) learning projections are not easy -> use GNN to automatically mix
% (2) not symmetric (exposure bias) -> symmetriclly
% (3) general adv -> conditional adv
Lately, video domain adaptation techniques~\cite{AMLS,TAN,TcoN} have emerged to address the domain shift in videos using adversarial learning. By segmenting the source and target videos into a set of fixed-length action clips, DAAA~\cite{AMLS} directly matches the segment representations from different domains with the 3D-CNN ~\cite{C3D} feature extractor. TA$^3$N~\cite{TAN} weights the source and target segments with a proposed temporal attention mechanism, forcing the model to attend the temporal features of low domain discrepancy. Different from the prior work that mainly concentrates on intra-domain interactions, TcoN~\cite{TcoN} proposes a cross-domain co-attention module to measure the affinity of the segment-pairs from source and target domains and further highlight the key segments shared by both domains. 

Nevertheless, existing adversarial video domain adaptation methodologies are limited in three aspects. First, when data distributions embody complex structures like videos, there is no guarantee for the two distributions to become sufficiently similar when the discriminator is fully confused, as illustrated in Figure \ref{fig:example}. Second, existing algorithms perform asymmetrically at the training and test stages. For instance, TcoN takes as input the source and target pairs and calculates the cross-domain attention scores at training stage, but inferences are done only based on the target data at the test time. This discrepancy unavoidably causes the exposure bias and deteriorates the model performance. Third, utilizing a general domain classifier for adversarial learning is only able to match marginal distributions~\cite{marginal}, and so does not align the class-conditional distributions~\cite{conditional,conditional1}. The video recognition models trained in this manner are hereby less likely to achieve the class-wise alignment. 

% (1) learning projections are not easy -> use GNN to automatically mix
% (2) not symmetric (exposure bias) -> symmetriclly
% (3) general adv -> conditional adv

\begin{figure*}[t]
    \centering
    \includegraphics[width=0.95\linewidth]{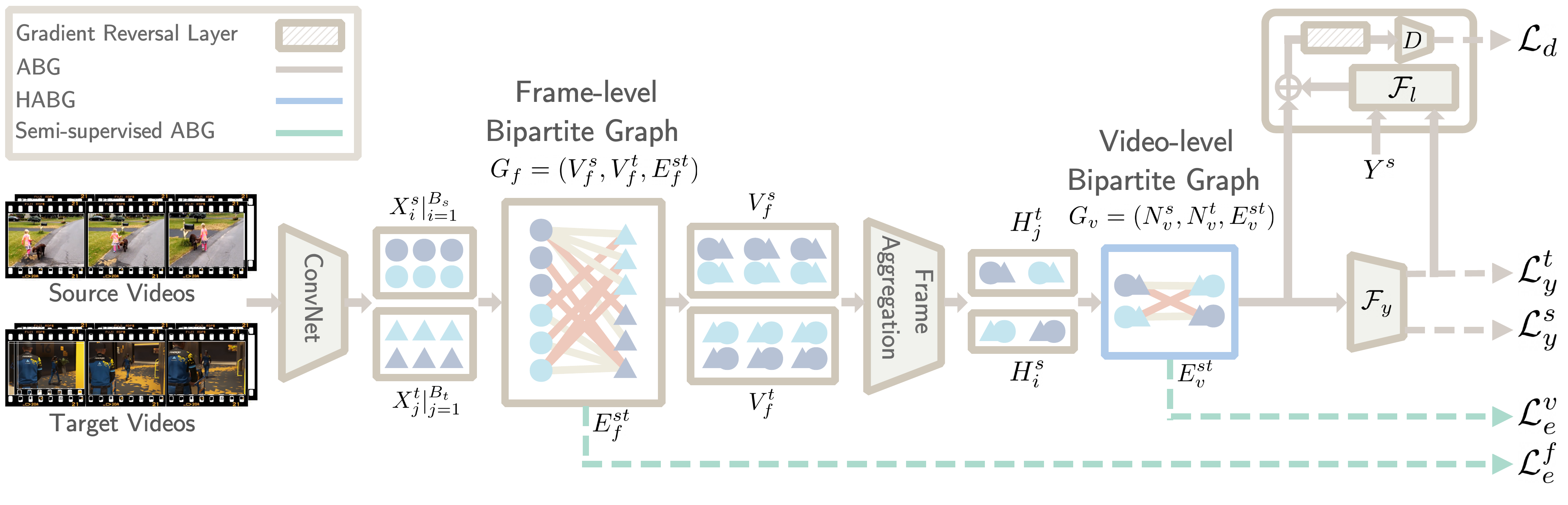}
    \caption{An overview of the proposed Adversarial Bipartite Graph (ABG) architecture, its hierarchical variant HABG (shown in blue), and the Semi-supervised ABG (shown in green).}
    \label{fig:flowchart}
\end{figure*}

To address the above-mentioned issues, in this paper, we take a more feasible strategy, \textit{i.e.}, to construct a domain-agnostic video classifier instead of pursuing with domain-invariant feature learning. In the proposed Adversarial Bipartite Graph (ABG) framework as illustrated in Figure \ref{fig:flowchart}, the video classifier is explicitly exposed to the mixed cross-domain representations, which preserves the temporal correlations across the domains modeled with a network topology of the bipartite graph. In particular, the source and target frames are sampled as heterogeneous vertexes of the bipartite graph, and the edges connecting the two types of nodes measure their similarity. Through message-passing, each vertex aggregates the features of its heterogeneous neighbors, making those from the similar source and target frames to be evenly mixed in the shared subspace. The proposed strategy performs symmetrically during the training and test phases, which successfully addresses the exposure bias issue.

Moreover, as the proposed framework is agnostic to the choices of frame aggregation, four different aggregation mechanisms are investigated, followed by a conditional adversarial module to preserve the class-specific consistency across the domains. The source labels and the target predictions are embedded as vectors which provide semantic cues for the domain classifier. To cope with large domain discrepancy, we additionally apply a video-level bipartite graph on the original model, called Hierarchical ABG. To testify the robustness of the proposed model, we further extend it to a semi-supervised domain adaptation setting (Semi-ABG), by adding the partial edge supervision. Extensive experiments conducted on four benchmark datasets evidence the superiority of the proposed adversarial bipartite framework over the state-of-the-art approaches. Overall, our contributions can be briefly summarized as follows:
\begin{itemize}
    \item We introduce a new Adversarial Bipartite Graph (ABG)  framework for unsupervised video domain adaptation, which focuses on recognizing domain-agnostic concepts rather than learning domain-invariant representations. It is further generalized to its hierarchical variant for challenging transfer tasks.
    \item To address the exposure bias issue, the proposed model is trained and tested symmetrically.
    \item The proposed ABG framework is seamlessly equipped with a conditional domain adversarial module which globally aligns the class-conditional distributions from different domains.
    \item We have demonstrated the effectiveness of the proposed strategy through extensive experiments on four large-scale video domain adaptation datasets and released the source code for reference. 
\end{itemize}

% The rest of paper is organized as follows. Section 2 presents a brief review of action recognition techniques and both image- and video-level domain adaptation approaches. Section 3 introduces the details of the proposed adversarial bipartite graph model. The experimental comparisons with state-of-the-art and ablation study are presented in Section 4, followed by the conclusions in Section 5.

%% file: 2_relatedwork.tex
\vspace{-0.7cm}
\section{Related Work}
\subsection{Video Action Recognition}
Activity recognition has been one of the core topics in computer vision areas, with a wide range of real-world applications including video surveillance~\cite{mm3}, environment monitoring and video captioning~\cite{videocaptioning,videocaptioning1,mm2,mm5}. A typical pipeline is leveraging a two-stream convolutional neural network to classify actions based on the individual video frames or local motion vectors~\cite{DBLP:conf/cvpr/KarpathyTSLSF14,DBLP:conf/nips/SimonyanZ14}. To better capture the action dynamics and gesture changes, later work models the long-term temporal information with recurrent neural networks~\cite{DBLP:journals/pami/DonahueHRVGSD17}, 3D convolutions~\cite{C3D}, and multi-scale temporal relation networks (TRN) ~\cite{TRN}. Another line of work augments the extracted RGB and optical flow features with multi-modal pose representations~\cite{pose}, complex object interactions~\cite{object}, and 3D human skeleton~\cite{skeleton,skeleton1}, which relieve the view dependency and the noises from different lighting conditions. However, the all above-mentioned work requires expensive annotations and could barely generalize to an unseen circumstance, which greatly hinders the feasibility in practice.

\subsection{Domain Adaptation}
Unsupervised Domain Adaptation (UDA) tackles such a limitation by trying to transfer knowledge from a labeled source domain to an unlabeled target domain. The discrepancy between the two domains refers to the domain shift~\cite{domain,domain1}, which is addressed by minimizing a distribution distance such as Maximum Mean Discrepancy (MMD)~\cite{MMD} with its variants~\cite{MMD1} and/or learning domain-invariant representations with adversarial learning~\cite{DANN} recently. Alternatively, an emerging line of work incorporates graph neural networks (GNN)~\cite{GCN,GraphSage,GAT} to bridge the domain gap at a manifold level, learning the intra-domain correlations in a transductive way. Very recently, GCAN~\cite{GCAN} constructs a densely-connected instance graph for the source and target nodes, and assigns pseudo labels for the target samples for aligning class centroids from different domains. While effective, existing graph-based work fails to model the inter-domain interactions, which makes it far from optimal.

\subsection{Video Domain Adaptation}
 Despite of the fact that domain adaptation has made great progress in a broader set of image recognition tasks, it is barely investigated for transferring knowledge across the videos. Early efforts~\cite{DBLP:journals/ivc/TangJTL16,DBLP:journals/ivc/XuZWF16} on video domain adaptation utilize a shallow model, that employs collective matrix factorization or PCA to learn a common latent semantic space for the source and target domains. Of late, the focus has shifted to the deep models~\cite{AMLS,TAN,TcoN}. Jamal \textit{et al.} projected the pre-extracted C3D \cite{C3D} representations of the source and target videos to a Grassmann manifold, and performed domain adaptation with adaptive kernels and adversarial learning. To extend the idea of modeling actions on the latent subspace, an end-to-end Deep Adversarial Action Adaptation (DAAA)~\cite{AMLS} is derived to learn the source and target video clips in the same temporal order. Zhang \textit{et al.} transferred from the trimmed video domain to the untrimmed video domain with Maximum Mean Discrepancy (MMD)~\cite{MMD} for action localization, yet without tasking into consideration any frame-level feature alignment. Chen \textit{et al.} proposed a temporal attentive adversarial adaptation network (TA$^3$N), which leverages the entropy of the domain predictor to attend the local temporal features of low domain discrepancy. Pan \textit{et al.}~\cite{TcoN} designed a co-attention module to minimize the domain discrepancy, which concentrates on the key segments shared by the both domains. Nevertheless, prior work is vulnerable and unreliable due to the overfitting and exposure bias issues. Instead, the proposed adversarial bipartite graph model is capable of learning domain-agnostic concepts and aligning class-conditional distributions locally and globally.

%% file: 3_method.tex
\section{Methodology}
In this section, we first formulate the task of unsupervised video domain adaptation, and then elaborate the details of the derived Adversarial Bipartite Graph (ABG) framework and its hierarchical variant (HABG). To testify the robustness of the proposed model, it is further generalized to a semi-supervised setting (Semi-ABG) in Section \ref{sec:semi}. 

\subsection{Problem Formulation}\label{sec:pf}
Give a labeled source video collection $\mathcal{D}_s = \{(X^s_i, y_i)\}_{i=1}^{N_s}$ and an unlabeled target video set $\mathcal{D}_t = \{X^t_j\}_{j=1}^{N_t}$ containing $N_s$ and $N_t$ videos respectively, the aim is to design a transfer network for predicting the labels $\mathcal{D}_t$ of the unlabeled target videos. The source and target domains are of different distributions yet they share the same label space $\mathcal{Y}\in\mathbb{R}^C$, where $C$ is the number of classes. Each source video $X^s_i$ or target video $X^t_j$ consists of $K$ frames, \textit{i.e.}, $X^s_i = \{x_{i}^{k}\}_{k=1}^{K}$ and $X^t_j = \{x_{j}^{k}\}_{k=1}^{K}$, where $x_i^k, x_j^k\in\mathbb{R}^{D}$ indicate the features of the $k$-th frame, and $D$ is the feature dimension of the vectors. For constructing each mini-batch, we forward $B_s$ source features and $B_t$ target features to update the proposed model.

\subsection{\hspace{-2ex} Adversarial Bipartite Graph Learning (ABG)}\label{sec:ABG}
To model the data affinity across two domains, it is natural to formulate the problem with a bipartite graph, whose vertices can be divided into two disjoint and independent sets, with the edges connecting the vertices from different sets. As discussed in Section \ref{sec:adv}, the general pipeline of ABG consists of (1) mixing the similar source and target features with the frame bipartite graph; (2) aggregating frame features into the global video representations; (3) aligning class-conditional distributions with adversarial learning; and (4) classifying the obtained source and target representations. To enhance the model capacity for difficult transfer tasks, we design a hierarchical structure HABG, incorporated in the video bipartite graph, as detailed in Section \ref{sec:video_bg}.

\hspace{-3ex}\textbf{Frame-level Bipartite Graph.}\label{sec:frame_bg}
Let the frame-level directed bipartite graph be $G_f = (V_f^s, V_f^t, E^{st}_f)$, where the cross-domain edge feature map $E^{st}_f\in\mathbb{R}^{K\cdot B_s\times K\cdot B_t}$ represents the node affinity between the pairs of source and target frames. The source vertex set $V_f^s = \{v^s_{ik}|_{k=1}^K\}_{i=1}^{B_s}\in\mathbb{R}^{K\cdot B_s\times D_v}$ and target vertex set $V_f^t = \{v^t_{jk}|_{k=1}^K\}_{j=1}^{B_t}\in\mathbb{R}^{K\cdot B_t\times D_v}$, with $D_v$ the vertex feature dimension, are expected to dynamically aggregate information across the domains based on the learned edge features, thus closing the domain gap gradually. The propagation rules for the cross-domain edge update and node update are elaborated as follows. 

\textit{Frame Edge Update.}
To calculate the similarity between the source and target frames, the normalized edge matrix is defined as,
\begin{equation}
    \begin{split}
        A^f &= \sigma\Big(\mathcal{F}_{fe}(|\mathcal{V}_f^s - \mathcal{V}_f^t|;\theta_{fe})\Big),\\
        \tilde{A}^f &= \frac{A^f_{i \cdot}}{\|A^f_{i \cdot}\|_1},
        E_f^{st} = \frac{\tilde{A}^f_{\cdot j}}{\|\tilde{A}^f_{\cdot j}\|_1},
    \end{split}
\end{equation}
with $\sigma$ the sigmoid function, and $\|\cdot\|_{1}$ the $L_1$ norm. $\mathcal{V}_f^s\in\mathbb{R}^{K\cdot B_s\times K\cdot B_t\times D_v}$ and $\mathcal{V}_f^t \in\mathbb{R}^{K\cdot B_s\times K \cdot B_t\times D_v}$ are the augmented tensors of the source and target vertexes, with the dimensions being expanded by repeating. $\mathcal{F}_{fe}(\cdot; \theta_{fe})$ is the frame-level metric network parameterized by $\theta_{fe}$, which computes the similarity scores between the source and target frames. To ease the impact of the number of cross-domain neighbors, the row normalization and column normalization are adopted on the edge feature map $E_{f}^{st}$.

\textit{Frame Node Update.} The generic rule to update node features can be formulated as follows,
\begin{equation}
    \begin{split}
        &\tilde{V}_f^s = E^{st}_f V_f^t ,~ \tilde{V}_f^t = (E^{st}_f)^{T} V_f^s,\\
        &V_f^s \leftarrow \mathcal{F}_{fv}([V_f^s; \tilde{V}_f^s];\theta_{fv}), V_f^t \leftarrow \mathcal{F}_{fv}([V_f^t; \tilde{V}_f^t];\theta_{fv}),
    \end{split}
\end{equation}
where $[\cdot;\cdot]$ is the concatenation operation, and $\mathcal{F}_{fv}(\cdot;\theta_{fv})$ is a node update network for both source and target nodes. The node embedding is initialized with the extracted representation from the backbone embedding model, \textit{i.e.}, $V_f^s=\{(x^1_i, x^2_i, \ldots, x^K_i)\}_{i=1}^{B_s}\in\mathbb{R}^{K\cdot B_s\times D}$, $V_f^t=\{(x^1_j, x^2_j, \ldots, x^K_j)\}_{j=1}^{B_t}\in\mathbb{R}^{K\cdot B_t\times D}$.

\hspace{-3ex}\textbf{Frame Aggregation.}\label{sec:agg}
To group the sampled frames into a unified video representation and capture appearance and temporal dynamics, the frame aggregation is applied on the learned source and target node embeddings. As the proposed framework is agnostic to the choices of frame aggregation, we examine multiple aggregation functions, including a symmetric average pooling function which is invariant to the order of frames, two memory based modules to capture the temporal information among frames, and a temporal relation network to explore the multi-scale temporal dynamics.

\textit{Mean Average Pooling.} By viewing the video as a collection of key frames, the video representation can be obtained by averaging the frame features temporally. Hence, each source video representation $H_i^s\in\mathbb{R}^{D_v}$ and target video representation $H_i^t\in\mathbb{R}^{D_v}$ are computed as,
\begin{equation}
    \begin{split}
        H^s_{i} = \frac{1}{K}\sum_{k=1}^K v^s_{ik}, H^t_{j} = \frac{1}{K}\sum_{k=1}^K v^t_{jk}.
    \end{split}
\end{equation}

\textit{Memory Based Aggregators.} Considering the temporal characteristics in human actions and events, two memory based aggregators, \textit{i.e.}, LSTM and GRU, are tested to construct the $i$-th source representation $H^s_{i} = H^s_{iK}\in\mathbb{R}^{D_v}$, and $j$-th target representation $H^t_{j} = H^t_{jK}\in\mathbb{R}^{D_v}$ as:
\begin{equation}
    \begin{split}
        H^s_{ik} = LSTM(H^s_{i k-1}, v^s_{ik})&, H^t_{jk} = LSTM(H^t_{j k-1}, v^t_{jk}),\\
        H^s_{ik} = GRU(H^s_{i k-1}, v^s_{ik})&, H^t_{jk} = GRU(H^t_{j k-1}, v^t_{jk}),
    \end{split}
\end{equation}
with $H$ the output hidden states, and $K$ the last step.

\textit{Temporal Relation Network (TRN).}
Inspired by \cite{TRN,TAN}, we further build up on a fine-grained relationship among the multi-scale video segments. In particular, the temporal relation network (TRN)~\cite{TRN} is able to preserve the short-term (\textit{e.g.}, 2-frame relation), and long-term (\textit{e.g.}, 5-frame relation) action dynamics, which potentially expands the temporal information that the learned video features could convey. The multi-scale temporal relations for the source data $H^s\in\mathbb{R}^{B_s\times D_v}$ and the target data $H^t\in\mathbb{R}^{B_t\times D_v}$ are defined as the composite functions below,

\begin{align}
        &T_2(V_f^s) = \mathcal{G}\Big(\sum_{k_1 < k_2}\mathcal{F}_{t}(v^s_{\cdot k_1}, v^s_{\cdot k_2})\Big), T_2(V_f^t) = \mathcal{G}\Big(\sum_{k_1 < k_2}\mathcal{F}_{t}(v^t_{\cdot k_1}, v^t_{\cdot k_2})\Big),\nonumber\\
        &H^s = \sum_{k=2}^KT_k(V_f^s), H^t = \sum_{k=2}^KT_k(V_f^t),
\end{align}
with the $T_2(\cdot)$ indicates the 2-frame local relation function. Note that the multi-scale function is the sum of the local relation scores from 2-frame to $K$-frame. The $\mathcal{F}_t(\cdot;\theta_t)$ and $\mathcal{G}(\cdot; \theta_G)$ are fully connected layers, fusing the features of different ordered frames.

\hspace{-3ex}\textbf{Video-level Bipartite Graph.}\label{sec:video_bg}
For difficult transfer tasks, we additionally apply a video-level bipartite graph on top of the frame aggregation network, fusing the source and target data hierarchically. It allows video features to be grouped into tighter clusters which improves classification performance. Similarly, we construct the video-level directed bipartite graph $G_v = (N_v^s, N_v^t, E_v^{st})$, where the $E_v^{st}\in\mathbb{R}^{B_s\times B_t}$ indicates the node affinity among the source and target videos. The source node set $N_v^s = \{n^s_{i}\}_{i=1}^{B_s}\in\mathbb{R}^{B_s\times D_n}$ and target node set $N_v^t = \{n^t_{j}\}_{j=1}^{B_t}\in\mathbb{R}^{B_t\times D_n}$, with $D_n$ the feature dimension of video nodes, are learned through message passing as defined below.

\textit{Video Edge Update.}
To calculate the similarity between the source and target frames, the normalized edge matrix is defined as,
\begin{equation}
    \begin{split}
        A^v &= \sigma\Big(\mathcal{F}_{ve}(|\mathcal{N}_v^s - \mathcal{N}_v^t|;\theta_{ve})\Big),\\
        \tilde{A}^v &= \frac{A^v_{i \cdot}}{\|A^v_{i \cdot}\|_1},
        E_v^{st} = \frac{\tilde{A}^v_{\cdot j}}{\|\tilde{A}^v_{\cdot j}\|_1},
    \end{split}
\end{equation}
with $\sigma$ the sigmoid function, $\|\cdot\|_{1}$ the $L_1$ norm. $\mathcal{N}_v^s\in\mathbb{R}^{B_s\times B_t\times D_n}$ and $\mathcal{N}_v^t\in\mathbb{R}^{B_s\times B_t\times D_n}$ are the augmented tensors of the source and target vertices, with the dimensions being expanded by repeating. $\mathcal{F}_{ve}(\cdot; \theta_{ve})$ is the video-level metric network parameterized by $\theta_{ve}$, computing the correlations among the source-target video pairs. 

\textit{Video Node Update.} The generic rule to update the node features can be formulated as follows,
\begin{equation}
    \begin{split}
        &\tilde{N}_v^s = E^{st}_v N^t_{v},~ \tilde{N}_v^t = (E^{st}_v)^T N^s_{v},\\
        &N_v^s \leftarrow \mathcal{F}_{vn}([N_v^s; \tilde{N}_v^s];\theta_{vn}), N_v^t \leftarrow \mathcal{F}_{vn}([N_v^t; \tilde{N}_v^t];\theta_{vn}),
    \end{split}
\end{equation}
where $[\cdot;\cdot]$ is the concatenation operation and $\mathcal{F}_{vn}(\cdot;\theta_{vn})$ is a node update network for the both source and target nodes. The node embeddings are initialized with the aggregated features \textit{i.e.}, $N_v^s=\{H_i^s\}_{i=1}^{B_s}\in\mathbb{R}^{B_s\times D_v}$, $N_v^t=\{H_j^t\}_{j=1}^{B_t}\in\mathbb{R}^{B_t\times D_v}$.

\hspace{-3ex}\textbf{Video Classification.}\label{sec:classify}
To predict the labels for the source and target samples, we construct a video classier $\mathcal{F}_y(\cdot;\theta_y)$ based on the aggregated video features for the ABG structure and the video vertex features for the HABG, respectively. Since the source data is labeled, the classifier is trained to minimize the negative log likelihood loss for each mini-batch,
\begin{align}
        \mathcal{L}^s_y = -\frac{1}{B_s}\sum_{i=1}^{B_s}y_i\log(\mathcal{F}_y(n_i^s)).
\end{align}
Instead of the supervised loss, for the unlabeled target data, a soft entropy based loss is adopted to alleviate the uncertainty of the predictions:
\begin{align}
        \mathcal{L}^t_y = -\frac{1}{B_t}\sum_{j=1}^{B_t}\mathcal{F}_y(n_j^t)\log(\mathcal{F}_y(n_j^t)).
\end{align}

\hspace{-3ex}\textbf{Conditional Adversarial Learning.}\label{sec:adv}
Besides leveraging bipartite graph neural networks to fuse the source and target neighbors, a conditional adversarial module is applied to align the class-conditional distributions. To achieve this, the module is composed of a label embedding function $\mathcal{F}_l(\cdot;\theta_l)$ and a domain classifier $\mathcal{D}(\cdot; \theta_d)$. The label embedding function projects the $i$-th source video label $y_i$ and the $j$-th target frame predictions $\mathcal{F}_y(n_{j}^t)$ into the latent vectors $\tilde{y}^s_i\in\mathbb{R}^{D_n}$ and $\tilde{y}^t_{j}\in\mathbb{R}^{D_n}$, providing the domain-invariant semantic cues for the domain classifier. The domain classifier is then conditioned on the classes for which the samples may belong to, and trained to discriminate between the features coming from the source or target data. The bipartite graphs are viewed as the feature generator to fool the discriminator. The adversarial objective function for the conditional adversarial module is formulated as:
\begin{align}
    &\mathcal{L}_d = \mathbb{E}_{n_i^s\sim N_v^s}\log[\mathcal{D}(n_i^s + \tilde{y}^s_i)] + \mathbb{E}_{n_j^t\sim N_v^t}\log[1 - \mathcal{D}(n_j^t + \tilde{y}_j^t)]\nonumber,\\
        &\tilde{y}_i^s = \mathcal{F}_l(y_i;\theta_l),~ \tilde{y}_j^t = \mathcal{F}_l(\mathcal{F}_y(n_j^t);\theta_l).
\end{align}
Consequently, the learned features will be more discriminative and aligned when the two-player mini-max game reaches an equilibrium. 
\vspace{-0.3cm}\subsection{Semi-supervised ABG and HABG}\label{sec:semi}
To verify the robustness of the proposed ABG and HABG structure, we further extend them to a semi-supervised setting. In this circumstance, part of the target labels $y^t_{\Omega}\in\mathbb{R}^{|\Omega|\times C}$ in a mini-batch are available for training. Here, we denote $\Omega$ and $\neg \Omega$ as the indices of the labeled and unlabeled target data, respectively. To fully take advantage of the partial target supervision, the classification objective functions are modified accordingly,
\begin{align}
     \mathcal{L}_y^s &= -\frac{1}{B_s}\sum_{i=1}^{B_s}y_i\log(\mathcal{F}_y(n_i^s)) - \frac{1}{|\Omega|}\sum_{j\in\Omega}y^t_j\log(\mathcal{F}_y(n_j^t)),\nonumber\\
        \mathcal{L}^t_y &= -\frac{1}{|\neg\Omega|}\sum_{j\in\neg\Omega}\mathcal{F}_y(n_j^t)\log(\mathcal{F}_y(n_j^t)),  
\end{align}
with $|\neg\Omega| = B_t - |\Omega|$. Moreover, the edge maps learned from either the frame-level or video-level bipartite graphs are able to be partially supervised. The newly added edge supervision is a binary cross entropy loss, which can be formulated as,
\begin{equation}
    \begin{split}
        \mathcal{L}_e^f &= \sum_{i=1,j\in\Omega}^{B_s}\sum_{k=1}^K E_f^{st}(i+k,j+k) \delta(y_i = y^t_j), \\
        \mathcal{L}_e^v &= \sum_{i=1,j\in\Omega}^{B_s} E_v^{st}(i,j) \delta(y_i = y^t_j),
    \end{split}
\end{equation}
where $\delta(\cdot)$ is the Kronecker delta function that is equal to one when $y_i = y^t_j$, and zero otherwise. $E_f^{st}(i+k,j+k)$ indicates the element from the $(i+k)$-th row and $(j+k)$-th column of the frame-level edge map, and $E_v^{st}(i,j)$ represents the element from the $i$-th row and $j$-th column of the video-level edge map.
\vspace{-0.3cm}\subsection{Optimization}
Our ultimate goal is to learn the optimal parameters for the proposed model,
\begin{equation}\label{eq:opt}
    \begin{split}
        &(\theta^*_{fe}, \theta^*_{fv}, \Theta^*_{a}, \theta^*_{ve}, \theta^*_{vn}, \theta^*_l, \theta^*_y) = \argmin \mathcal{L} - \beta \mathcal{L}_d,\\
        &\theta_{d}^* = \argmin \mathcal{L} + \beta\mathcal{L}_d,\\
        &\mathcal{L} = \mathcal{L}_{y}^s + \gamma\mathcal{L}_{y}^t + \lambda (\mathcal{L}_e^v + \alpha\mathcal{L}_e^f),
    \end{split}
\end{equation}
with $\Theta_a$ being the learnable parameters of the frame aggregation module, and $\beta$, $\gamma$, $\lambda$ and $\alpha$ being the loss coefficients respectively. Notably, for the UDA setting, we have, $\mathcal{L}_{e}^f = 0$, and $\mathcal{L}_e^v = 0$. The overall algorithm is provided in supplementary material.

%% file: 4_experiments.tex
\begin{table*} 
\centering 
\caption{The general statistics of the four datasets used in our experiments.} \vspace{-0.3cm}% Table caption, can be commented out if no caption is required
\begin{tabular}{l | c c c c } % The final bracket specifies the number of columns in the table along with left and right borders which are specified using vertical pipes (|); each column can be left, right or center-justified using l, r or c. Columns will widen to hold the content in them by default, to specify a precise width, use p{width}, \textit{e.g.} p{5cm}
\toprule % Top horizontal line
% Amalgamating several columns into one cell is done using the \multicolumn command with the number of columns to amalgamate as the first argument and then the justification (l, r or c)
% Horizontal line spanning less than the full width of the table - you can add (r) or (l) just before the opening curly bracket to shorten the rule on the left or right side
\textbf{Property} &\textbf{UCF-HMDB$_{small}$} &\textbf{UCF-HMDB$_{full}$} &\textbf{UCF-Olympic} &\textbf{Kinetics-Gameplay} \\ % Column names row
\midrule % In-table horizontal line
Video Length & $\sim$21 Seconds & $\sim$33 Seconds &$\sim$39 Seconds &$\sim$ 10 Seconds\\ % Content row 1
Classes &5 &12 &6 &30\\ % Content row 2
Training Videos &UCF: 482 / HMDB: 350 &UCF:1,438 / HMDB: 840 &UCF: 601 / Olympic: 250 & Kinetics: 43,378 / Gameplay: 2,625 \\ % Content row 3
Validation Videos &UCF: 189 / HMDB: 571 &UCF: 360 / HMDB: 350 &UCF: 240 / Olympic: 54 & Kinetics: 3,246 / Gameplay: 749 \\ % Content row 4
\bottomrule % Bottom horizontal line
\end{tabular}
\label{tab:datasets}\vspace{-0.3cm}
\end{table*}
\vspace{-0.2cm}
\section{Experiments}
\subsection{Datasets}
We compare and contrast our proposed approach with the existing domain adaptation approaches on four benchmark datasets, \textit{i.e.}, the \textbf{UCF-HMDB$_{small}$}, \textbf{UCF-HMDB$_{full}$}, \textbf{UCF-Olympic} and \textbf{Kinetics-Gameplay}. For fair comparison, we follow the dataset partition and feature extraction strategies from \cite{TAN}, that utilizes the ResNet101 model pre-trained on ImageNet as the frame-level feature extractor. The statistics of the four datasets are summarized in Table \ref{tab:datasets}. The \textbf{UCF-HMDB$_{small}$} and \textbf{UCF-HMDB$_{full}$} are the overlapped subsets of two large-scale action recognition datasets, \textit{i.e.}, the UCF101~\cite{ucf} and HMDB51~\cite{hmdb}, covering 5 and 12 highly relevant categories respectively. The \textbf{UCF-Olympic} selects the shared 6 classes from the UCF101 and Olympic Sports Datasets~\cite{olympic}, including \textit{Basketball, Clearn and Jerk, Diving, Pole Vault, Tennis and Discus Throw}. The \textbf{Kinetics-Gameplay} is the most challenging cross-domain dataset, with a large domain gap between the synthetic videos and real-world videos. The dataset is build by selecting 30 shared categories between Gameplay~\cite{TAN} and one of the largest public video datasets Kinetics-600~\cite{kinetics}. Each category may also correspond to multiple categories in both dataset, which poses another challenge of class imbalance.
\vspace{-0.3cm}

\subsection{Baselines}
We compare our approach with several state-of-the-art video domain adaptation methods, image domain adaptation approaches, single-domain action recognition models, and a basic ResNet-101 classification model pre-trained on the ImageNet dataset. Single-domain action recognition models include the 3D ConvNets (\textbf{C3D}) \cite{C3D} and Temporal Segment Networks (\textbf{TSN})~\cite{TSN}, which are pre-trained on the source domain and tested on the target domain. Four classical image-level domain adaptation methods, \textit{i.e.}, Domain-Adversarial Neural Network (\textbf{DANN}) \cite{DANN}, Joint Adaptation Network (\textbf{JAN}) \cite{JAN}, Adaptive Batch Normalization (\textbf{AdaBN}) \cite{AdaBN} and Maximum Classifier Discrepancy (\textbf{MCD}) \cite{MCD} are adjusted to align the distributions of video features with the frame aggregation module. As for non-deep video domain adaptation, we compare the proposed \textbf{HABG} method with \textbf{Many-to-One}~\cite{DBLP:journals/ivc/XuZWF16} Encoder, two variants of Action Modeling on Latent Subspace (\textbf{AMLS})~\cite{AMLS}, \textit{i.e.}, the Subspace Alignment (\textbf{AMLS-SA}) and Geodesic Flow Kernel (\textbf{AMLS-GFK}). For deep video domain adaptation methods, we adopt the Deep Adversarial Action Adaptation (\textbf{DAAA})~\cite{AMLS}, Temporal Adversarial Adaptation Network (\textbf{TA$^{2}$N})~\cite{TAN}, Temporal Attentive Adversarial Adaptation Network (\textbf{TA$^{3}$N})~\cite{TAN} and Temporal Co-attention Network (\textbf{TCoN})~\cite{TcoN} for comparison.

\subsection{Implementation Details} % Backbone Model /
Our source code is based on PyTorch \cite{pytorch}, which is available in a Githu repository\footnote{https://github.com/Luoyadan/MM2020\_ABG} for reference. All experiments are conducted on two servers with two GeForce GTX 2080 Ti GPUs.

\subsubsection{Video Pre-processing}~\label{dp}
Following the standard protocol used in~\cite{TAN}, we sample a fixed-number $K$ of frames with an equal spacing from each video for training, and encode each frame with the Resnet-101 \cite{resnet} pre-trained on ImageNet into a 2048-D vector, \textit{i.e.}, $D=2048$. For fair comparison, we set $K$ to 5 in our experiments.

\begin{figure}[t]
    \centering
    \subfloat[][HMDB$\rightarrow$UCF$_{small}$]{\includegraphics[width=0.49\linewidth]{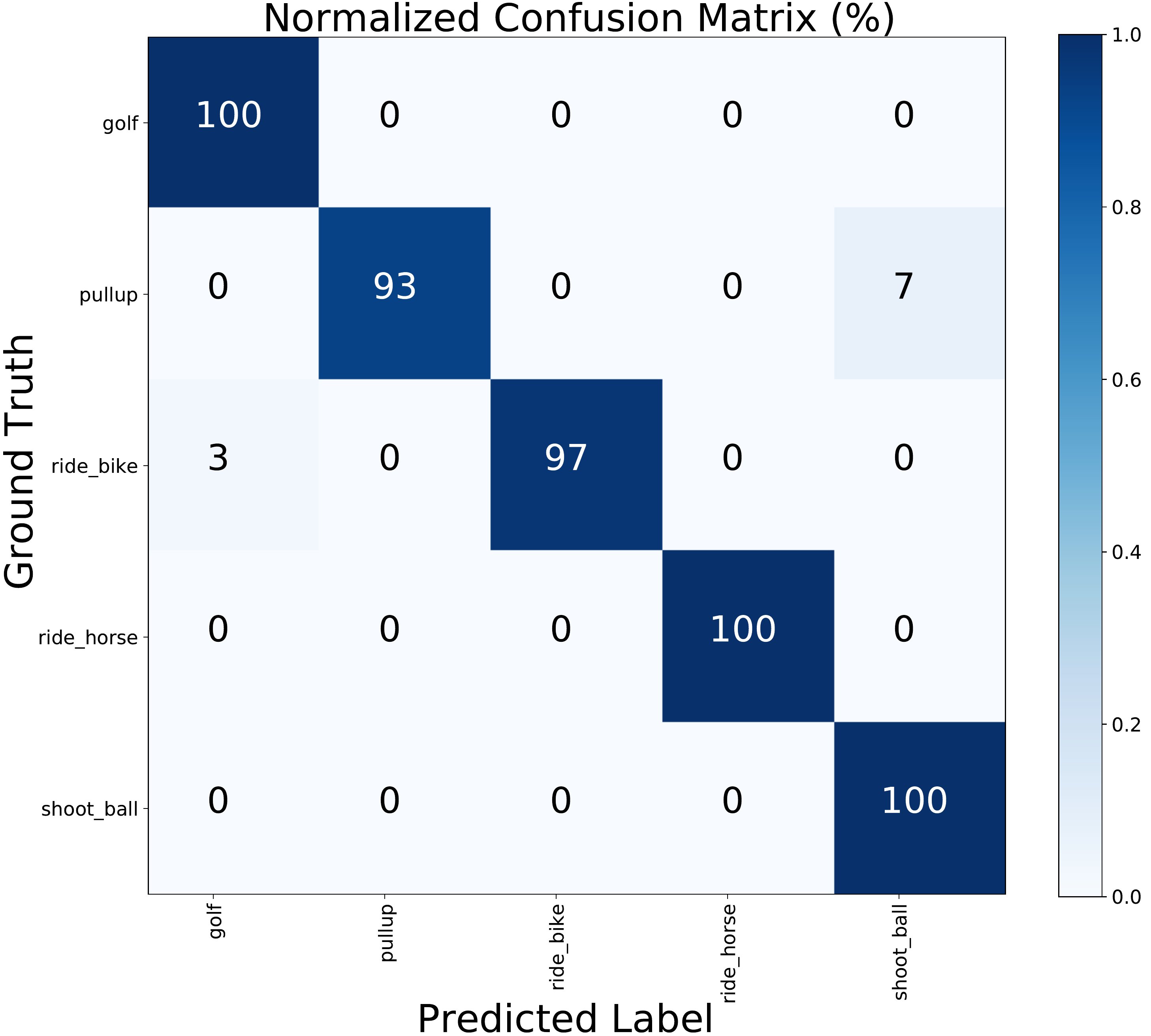}}
    \subfloat[][UCF$\rightarrow$HMDB$_{small}$]{\includegraphics[width=0.49\linewidth]{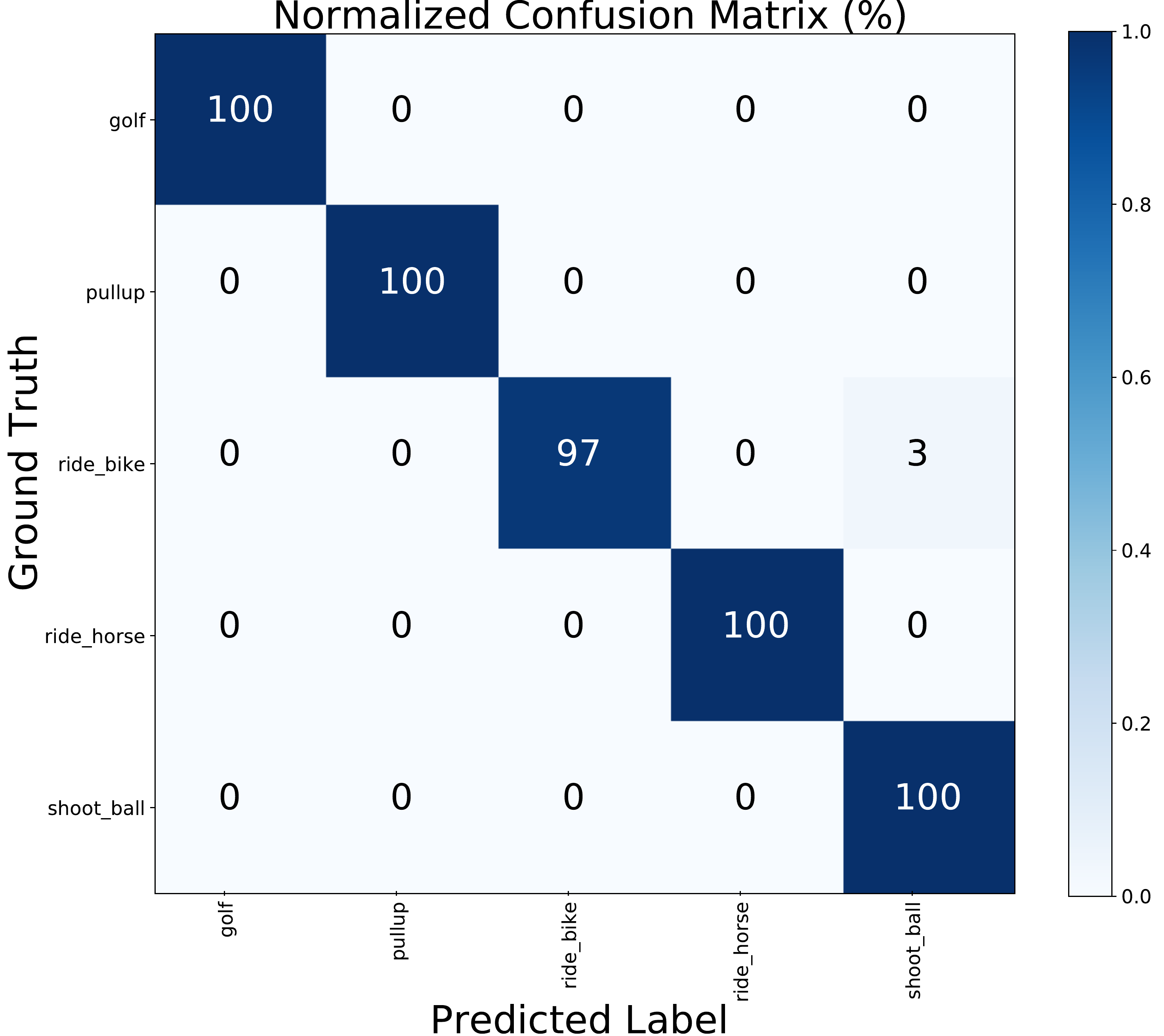}}\\
    \subfloat[][HMDB$\rightarrow$UCF$_{full}$]{\includegraphics[width=0.49\linewidth]{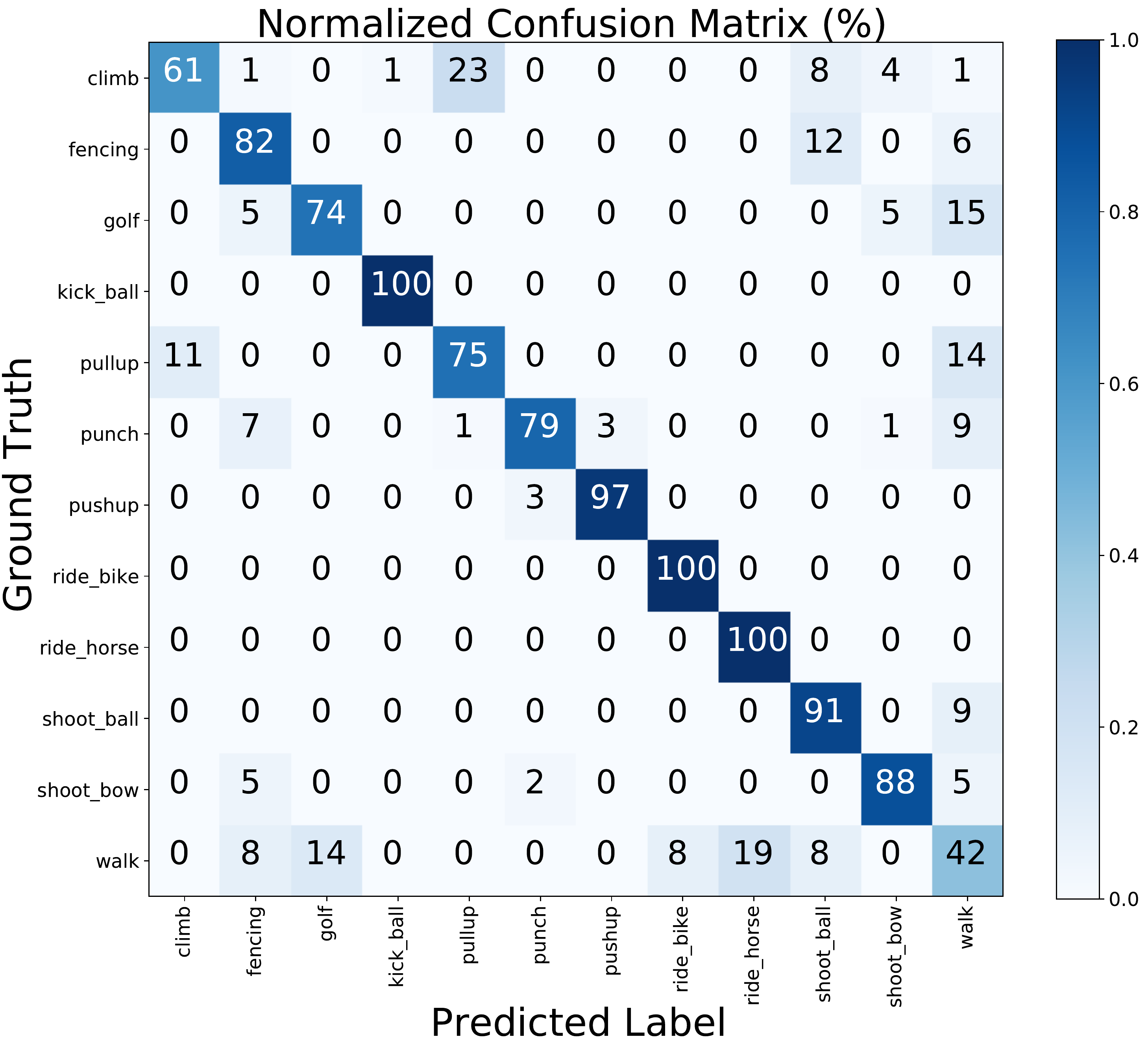}}
    \subfloat[][UCF$\rightarrow$Olympic]{\includegraphics[width=0.49\linewidth]{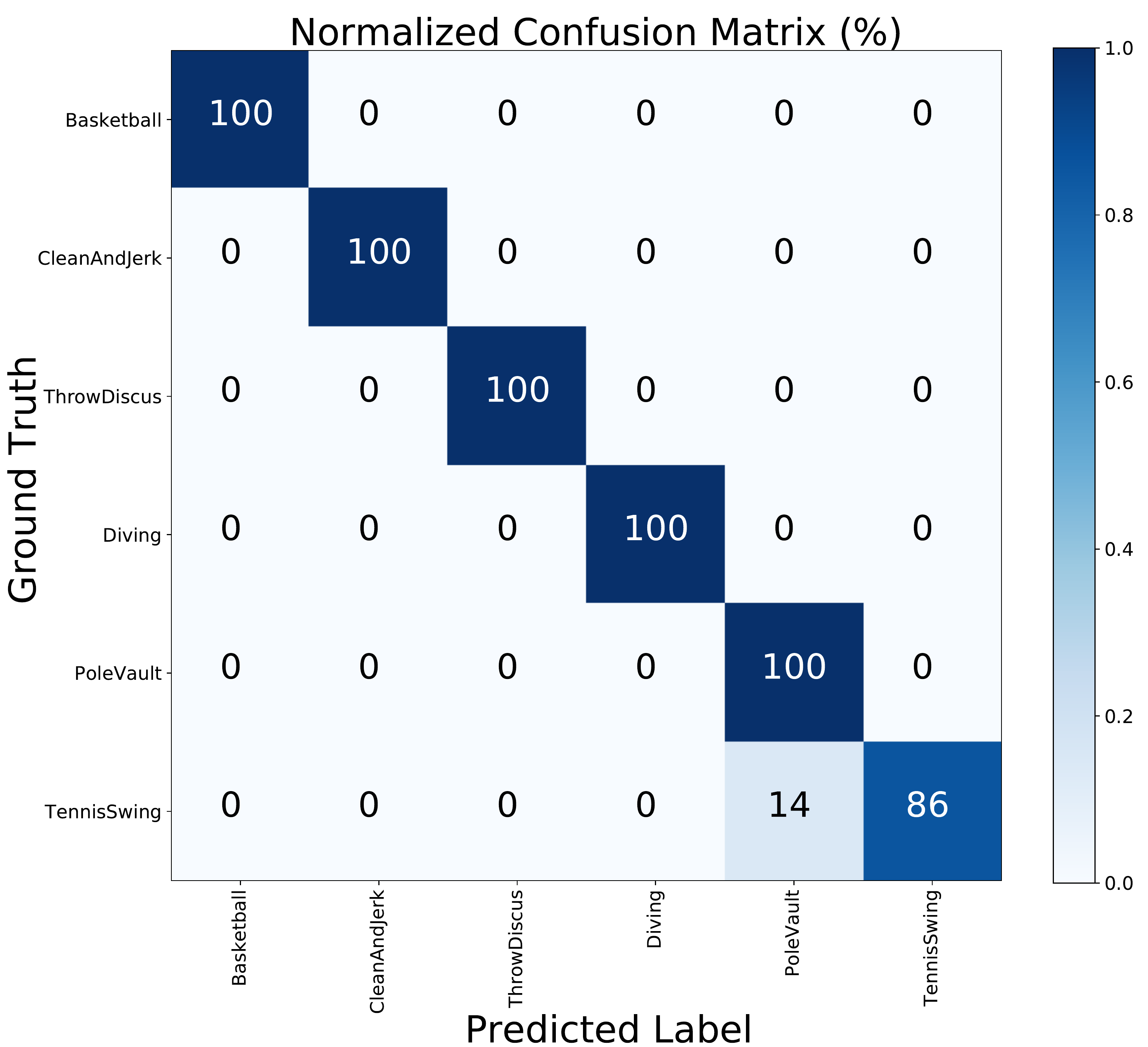}}
    \caption{The confusion matrices of the proposed ABG method performed on three benchmark datasets.}
    \label{fig:cf}\vspace{-0.7cm}
\end{figure}

\subsubsection{Module Architecture.}
The edge update networks $\mathcal{F}_{fe}$ and $\mathcal{F}_{ve}$ project the affinity map from $D$ and $D_v$ dimensions to 1-dim scores, which are composed of the two convolutional layers, batch normalization, LeakyReLU and edge dropout. The node update networks $\mathcal{F}_{fv}$ and $\mathcal{F}_{vn}$ map the $2D$-dim and $2D_v$-dim concatenation of node features and the neighbors' features to the $D_v$-dim and $D_n$-dim vectors, respectively. $\mathcal{F}_{fv}$ and $\mathcal{F}_{vn}$ include two convolutional layers, batch normalization, LeakyReLU and node dropout.

\subsubsection{Parameter Settings.}
 The hidden size, the feature dimension of frame nodes (\textit{i.e.} $D_v$) and the feature dimension of video nodes (\textit{i.e.} $D_n$) are fixed to 512. The total number of training epochs $M$ is 60 for Kinetics-Gameplay dataset, and 30 for the rest of datasets. The batch size $B_s$, $B_t$ for the source data and target data are set to 128. The stochastic gradient optimizer (SGD) is used as the optimizer with a momentum of 0.9 and weight decay of $1\times10^{-4}$. The learning rate $\mu$ is initiated as $4\times10^{-2}$ then decayed as the number of epoch increases, which follows the rule used in \cite{DANN,TAN}. The loss coefficients $\alpha$ and $\lambda$ are empirically fixed at 0.1 and 1 for semi-supervised experiments. The dropout rate is set to 0.2.

\subsection{Comparisons with State-of-The-Art}
 Under the unsupervised domain adaptation protocol, we compare the proposed ABG method with multiple baseline approaches on \textbf{UCF-HMDB$_{small}$}, \textbf{UCF-Olympic} and \textbf{UCF-HMDB$_{full}$} datasets. With different backbone networks, the comparison results achieved from the relatively small datasets are reported in Table \ref{tab:toy}. Table \ref{tab:UCF-HMDB} presents the results on the full UCF-HMDB dataset using various frame aggregation strategies. It is observed that the proposed \textbf{ABG} framework is superior to all the compared image- and video-level domain adaptation methods in most cases, especially achieving a significant performance boost on the large-scale testbed. Notably, \textbf{Source Only} indicates the backbone model pretrained on the source domain and tested on the target domain. \textbf{Target Only} denotes the backbone model trained and tested on the target domain. From Table \ref{tab:toy}, it is demonstrated that the deep video DA methods (line 6-10) generally outperforms the non-deep video DA approaches (line 3-5) and the classification models without DA (line 1-2). Among the deep \textbf{TCoN} and \textbf{TA$^3$N} leverage the attention mechanism and then suppress the variance caused by the outlier frames, improving the recognition accuracy by up to $6.3\%$ and $11.1\%$ over \textbf{DAAA} on the UCF-Olympic dataset. With the same backbone of TA$^3$N, our \textbf{ABG} model performs comparably without relying on the frame attention or complex frame aggregation strategies. This phenomenon is also observed in the large \textbf{UCF-HMDB$_{full}$} dataset, in which the proposed \textbf{ABG} with the average pooling boosts the performance by up to $10.1\%$ and $11.5\%$ over the state-of-the-art \textbf{TA$^3$N} on the UCF$\rightarrow$HMDB and HMDB$\rightarrow$UCF tasks, respectively. As shown in Table \ref{tab:UCF-HMDB}, average pooling (\textbf{AvgPool}) and \textbf{LSTM} suits the proposed model better among other frame aggregation functions. We infer the reasons behind is the frame bipartite graph has already fused similar frames regardless the order, which weakens the power of multi-scaled \textbf{TRN} aggregation. Notably, it is observed that \textbf{AdaBN} surpasses the most of image domain adaptation methods. As it separates the batch normalization layer for source and target data, AdaBN minimizes the risk of being overfitting to the source domain, which provides a strong support to our statement discussed in Section \ref{sec:intro}. To further investigate the detailed performance of the proposed ABG with respect to specific classes, four confusion matrices are provided in Figure \ref{fig:cf}.

\begin{table}[t]
	\caption{Recognition accuracies (\%) on the UCF-HMDB$_{small}$ and UCF-Olympic datasets. U: UCF, H: HMDB, O: Olympic.}\vspace{-0.2cm}
	\label{tab:toy}
	\centering
	\resizebox{1\linewidth}{!}{% 
		\begin{tabular}{l ccccc}
			\toprule
			\textbf{Method} &\textbf{Backbone} &\textbf{U$\rightarrow$H$_{small}$}	&\textbf{H$\rightarrow$U$_{small}$}	& \textbf{U$\rightarrow$O} & \textbf{O$\rightarrow$U}\\
			\midrule
			\multirow{2}{*}{Source Only} 
			&TSN &- &82.10 &80.00 &76.67\\
			&C3D &- &- &82.13 &83.16\\
			\midrule
			Many-to-One \cite{DBLP:journals/ivc/XuZWF16} &Action Bank &82.00 &82.00 &87.00 &75.00\\
			AMLS-SA \cite{AMLS} &C3D &90.25 &94.40 &83.92 &86.07\\
			AMLS-GFK \cite{AMLS} &C3D &89.53 &95.36 &84.65 & 86.44\\
			\midrule
			DAAA \cite{AMLS} &TSN &- &88.36 &88.37 &86.25\\
			DAAA \cite{AMLS} &C3D &- &- &91.60 &89.96\\
			TCoN \cite{TcoN} &TSN & - &93.01  &93.91 &91.65\\
			TA$^{3}$N \cite{TAN} &ResNet-101 &99.33 &\textbf{99.47} &\textbf{98.15} &\textbf{92.92}\\
			\midrule
			\textbf{ABG-AvgPool}  &ResNet-101 &\textbf{99.33} &98.41 &\textbf{98.15} &92.50\\
			\bottomrule
		\end{tabular}
	}\vspace{-0.3cm}
\end{table}

\begin{table}[t] % Add the following just after the closing bracket on this line to specify a position for the table on the page: [h], [t], [b] or [p] - these mean: here, top, bottom and on a separate page, respectively
	\centering % Centres the table on the page, comment out to left-justify
	\caption{Recognition accuracies (\%) of the domain adaptation methods and the proposed ABG model with respect to various frame aggregation strategies on the full UCF-HMDB dataset.}\vspace{-0.2cm}
	\resizebox{1\linewidth}{!}{% 
		\begin{tabular}{l c c c c c c c c}
			\toprule % Top horizontal line
			& \multicolumn{4}{c}{\textbf{UCF$\rightarrow$HMDB}} & \multicolumn{4}{c}{\textbf{HMDB$\rightarrow$UCF}}\\ 
			\cmidrule(l){2-5}\cmidrule(l){6-9} 
			\textbf{Method} &AvgPool &LSTM &GRU &TRN &AvgPool &LSTM &GRU &TRN\\ 
			\midrule % In-table horizontal line
			Source Only &70.28 &69.17 &70.83 &71.67 &74.96 &70.05 &76.36 &73.91\\ 
			\midrule
			DANN~\cite{DANN} &71.11 &70.00 &70.83 &75.28 &75.13 &75.83 &75.13 &76.36\\ 
			JAN~\cite{JAN} &71.39 &70.56  &72.50  &74.72 &77.58 &77.58 &77.75 &79.36\\
			AdaBN~\cite{AdaBN} &75.56 &74.17  &74.72  &72.22 &76.36 &77.41 &74.96 &77.41\\
			MCD~\cite{MCD} &71.67 &70.00 &74.44 &73.89 &76.18 &68.30 &78.81 &79.34\\ 
			\midrule % In-table horizontal line
			TA$^{2}$N~\cite{TAN}  &71.11 &70.00  &70.83 &77.22 &76.36 &70.75 &76.89 &80.56\\
			TA$^{3}$N~\cite{TAN}  &71.94 &70.00  &69.72 &\textbf{78.33} &76.36 &70.75 &77.23 &81.79\\
			\midrule % In-table horizontal line
			\textbf{ABG} &\textbf{79.17} &\textbf{75.56}  &\textbf{75.56}  &76.67  &\textbf{85.11} &\textbf{84.24} &\textbf{83.36} &\textbf{81.79}\\ 
			\midrule
			Target Only &80.56 &- &- &82.78 &92.12 &- &-  &94.92\\
			\bottomrule \vspace{-1cm}
		\end{tabular}
	}
	\label{tab:UCF-HMDB} 
\end{table}

\begin{figure*}[!htb]
    \centering\vspace{-2.2cm}
    \subfloat[][DANN]{\includegraphics[width=0.23\linewidth, height=3.5cm]{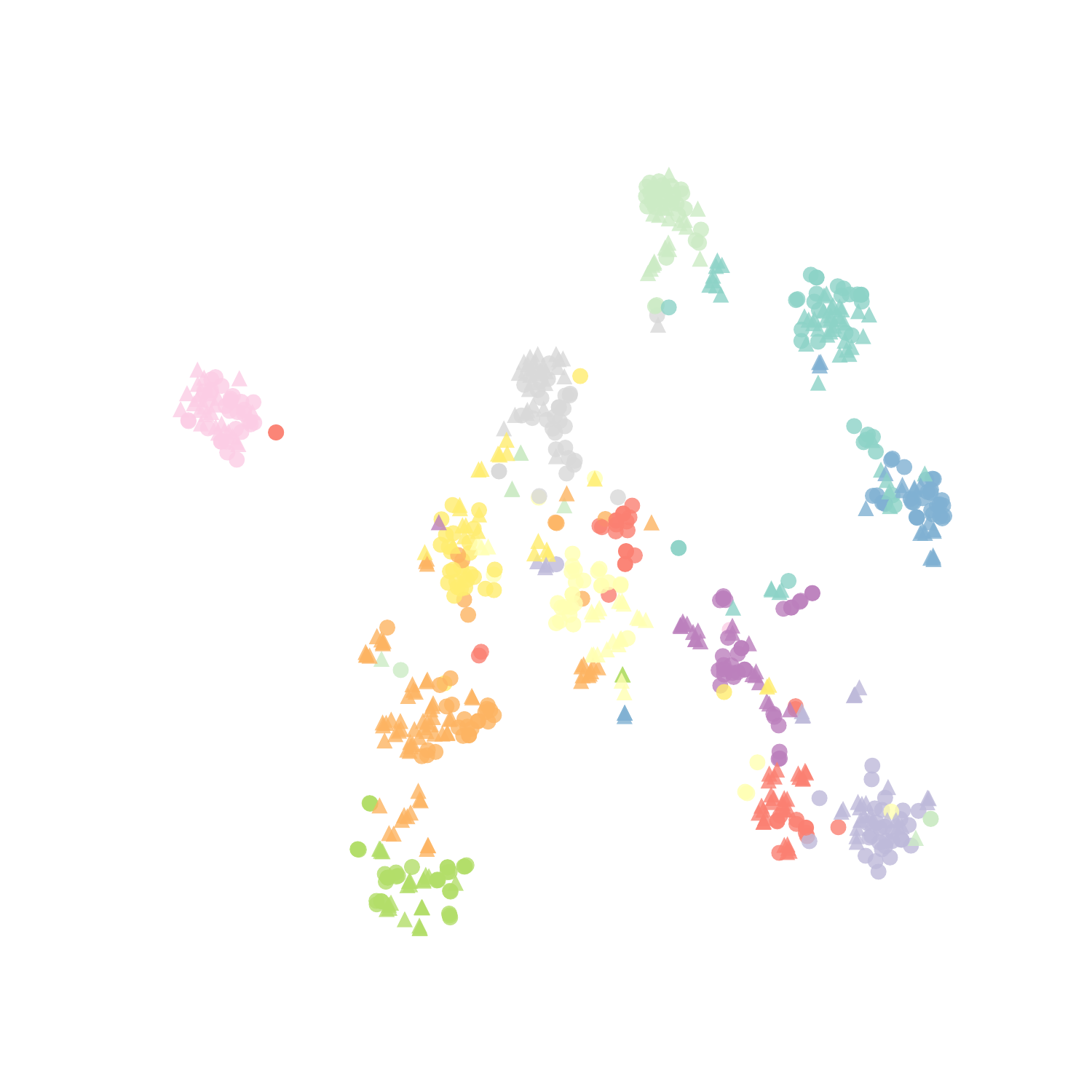}}
    \subfloat[][JAN]{\includegraphics[width=0.23\linewidth,height=3.5cm]{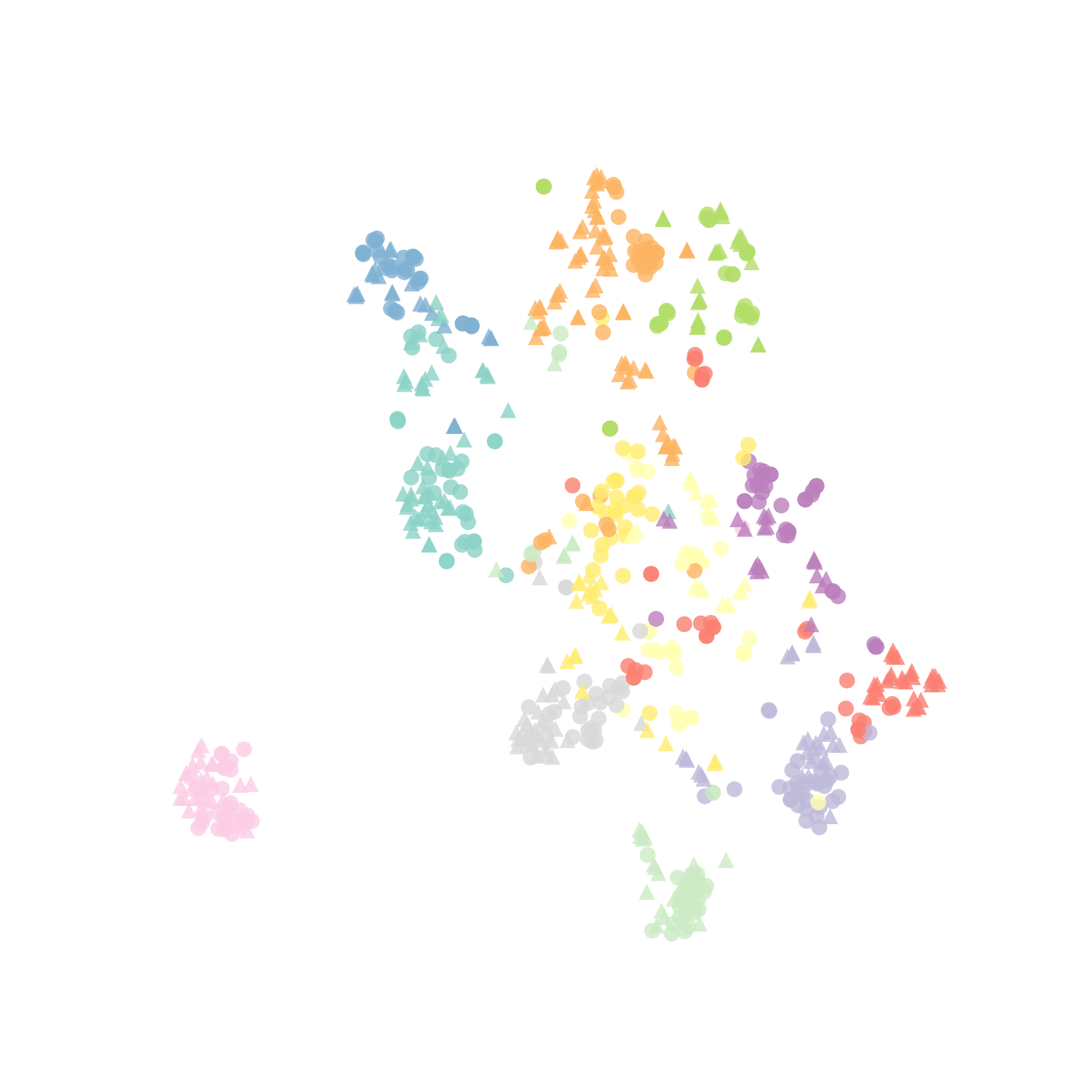}}
    \subfloat[][MCD]{\includegraphics[width=0.23\linewidth,height=3.5cm]{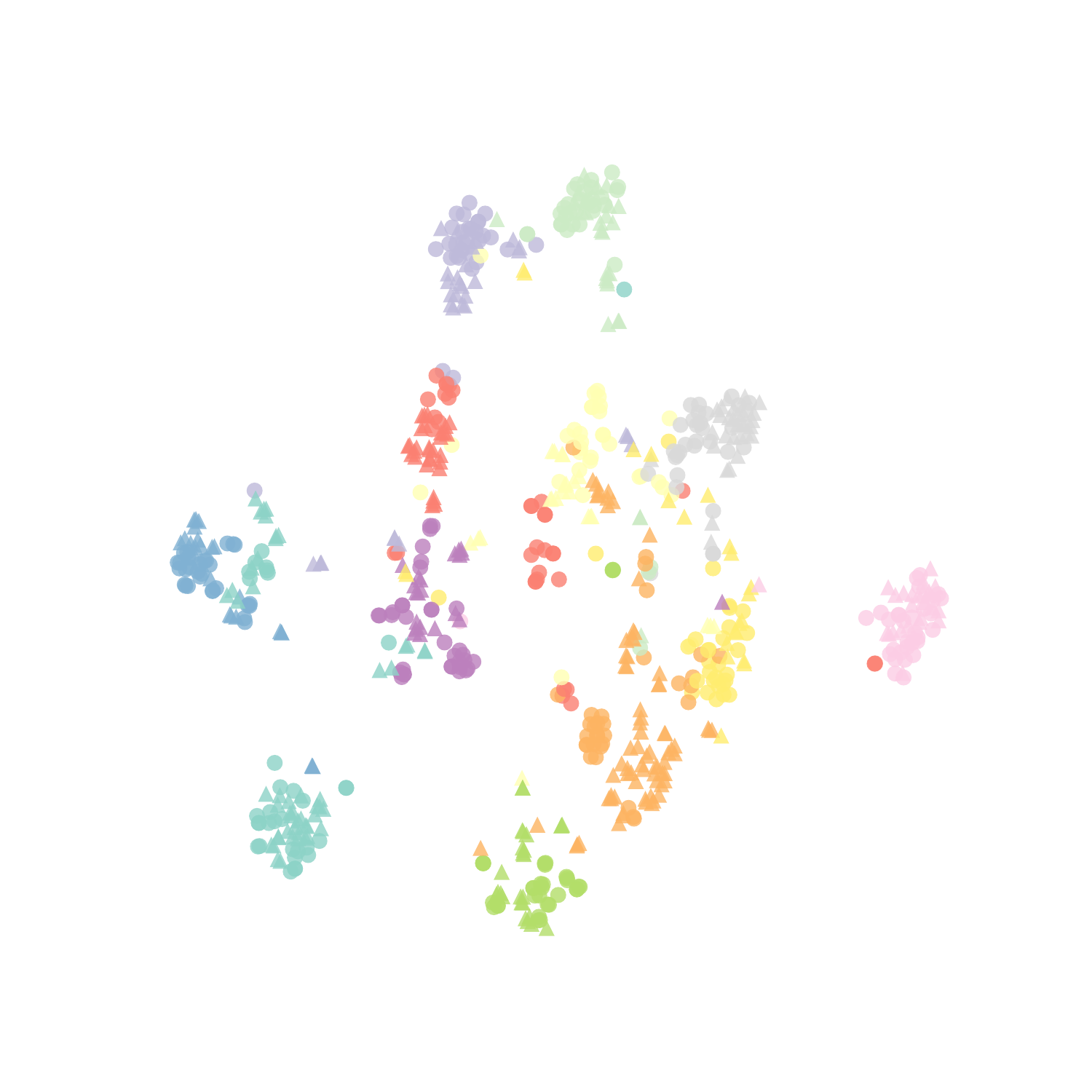}}
    \subfloat[][TA$^{3}$N]{\includegraphics[width=0.23\linewidth,height=3.5cm]{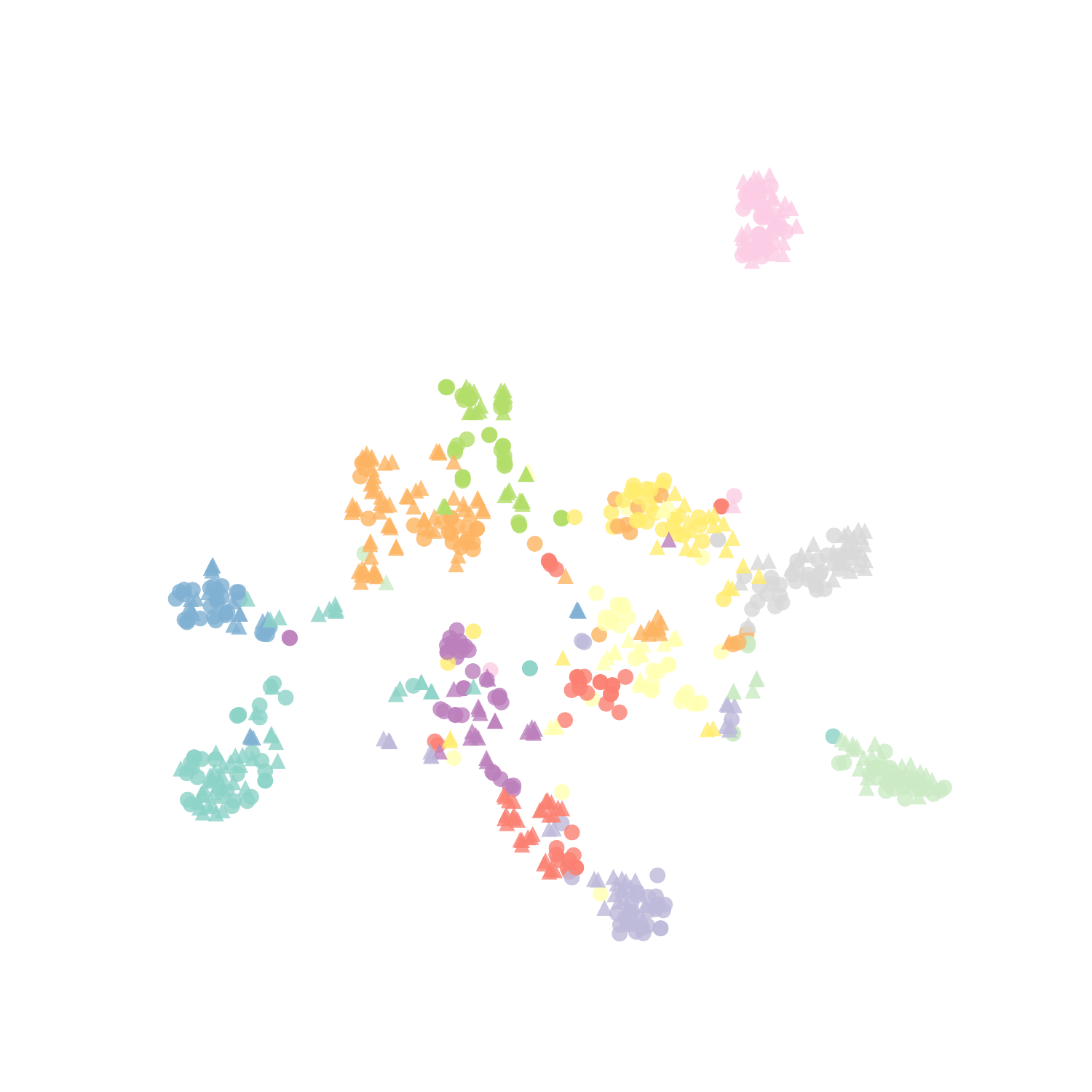}}\\\vspace{-0.3cm}
    \subfloat[][ABG-AvgPool]{\includegraphics[width=0.23\linewidth,height=3.5cm]{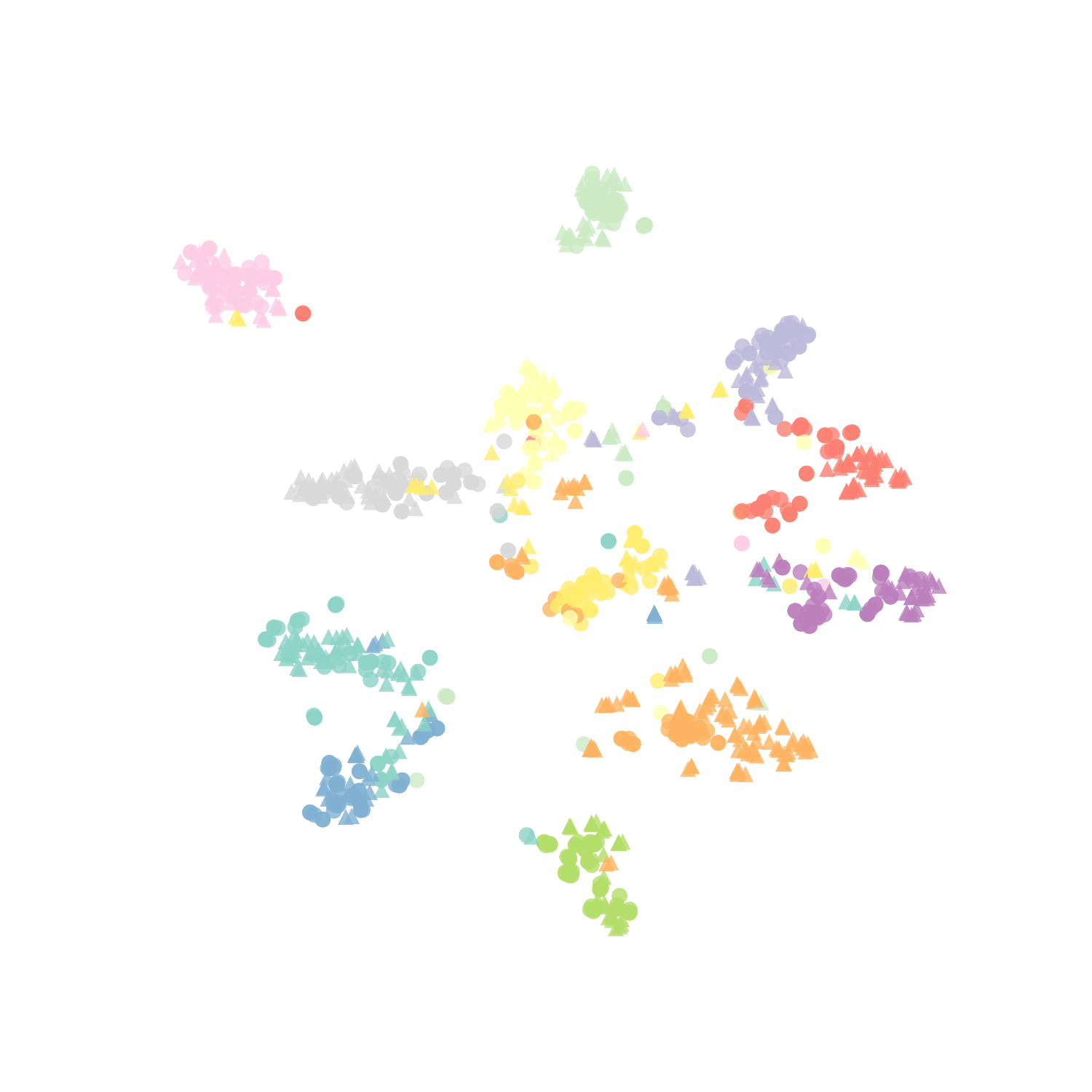}}
    \subfloat[][ABG-LSTM]{ \includegraphics[width=0.23\linewidth,height=3.5cm]{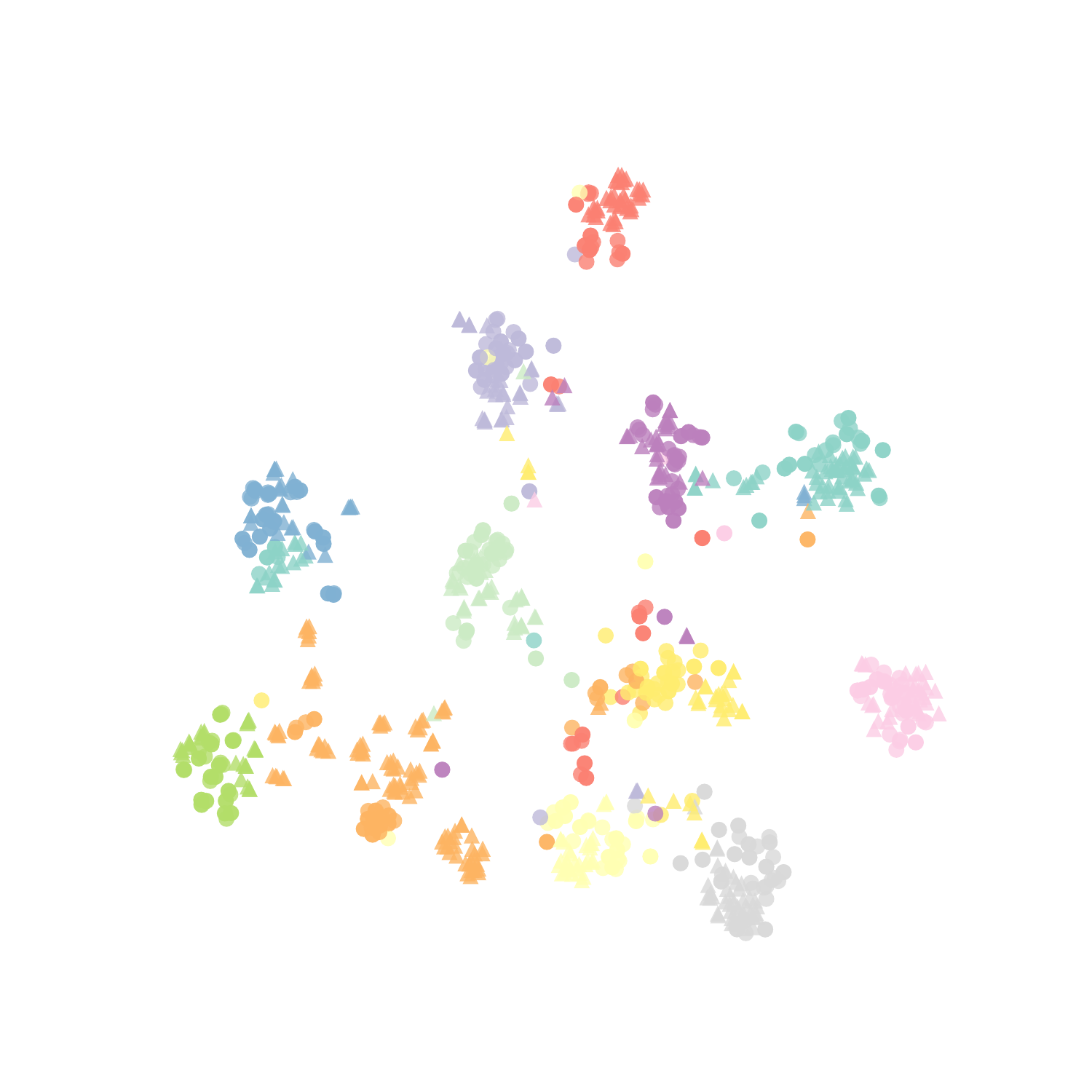}}
     \subfloat[][ABG-GRU]{\includegraphics[width=0.23\linewidth,height=3.5cm]{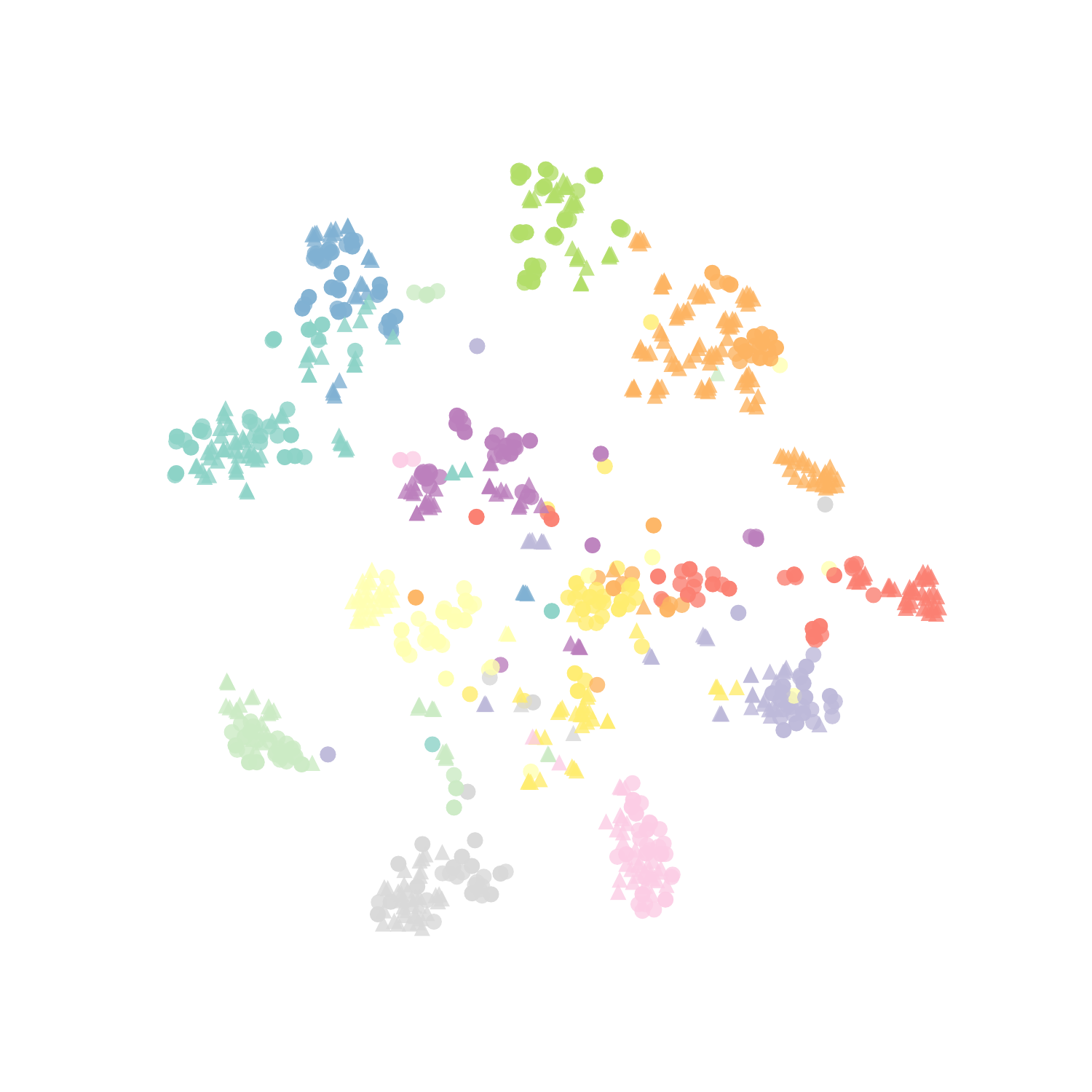}}
     \subfloat[][ABG-TRN]{\includegraphics[width=0.23\linewidth,height=3.5cm]{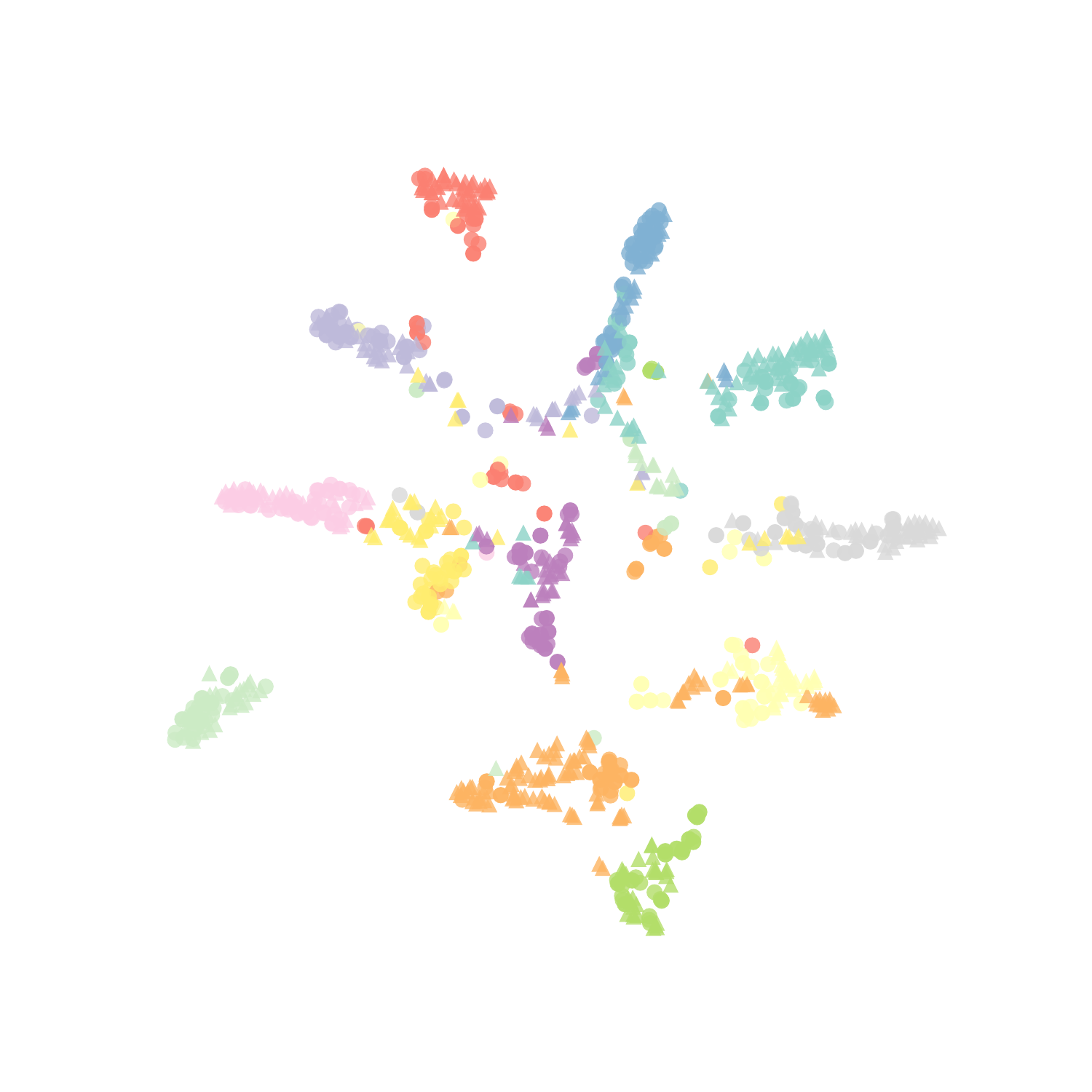}}
    \vspace{-0.2cm}\caption{The t-SNE visualization of the learned source and target video representations on the HMDB$\rightarrow$UCF task.}
    \label{fig:tsne}
\end{figure*}

\begin{table}[t]
\caption{Experimental results under the semi-supervised setting on the Kinetics-Gameplay dataset.}\vspace{-0.2cm}
\resizebox{1\linewidth}{!}{% 
\begin{tabular}{l c c c c c c c c} 
\toprule % Top horizontal line
& \multicolumn{4}{c}{\textbf{Kinetics$\rightarrow$Gameplay}} & \multicolumn{4}{c}{\textbf{Gameplay$\rightarrow$Kinetics}}\\ 
\cmidrule(l){2-5}\cmidrule(l){6-9} % Horizontal line spanning less than the full width of the table - you can add (r) or (l) just before the opening curly bracket to shorten the rule on the left or right side
\textbf{Method} &30\% &50\% &70\% &90\% &30\% &50\% &70\% &90\%\\ % Column names row
\midrule % In-table horizontal line
Source Only &16.29 &16.29 &16.29 &16.29 &14.82 &14.82 &14.82 &14.82\\ % Content row 1
\midrule
DANN~\cite{DANN} &49.67 &56.74 &60.48 &65.02 &33.24 &40.05 &40.44 &43.38\\ % Content row 2
JAN~\cite{JAN} &47.40 &53.94 &60.35 &61.28 &32.69 &22.95 &41.31 &32.75\\ % Content row 3
AdaBN~\cite{AdaBN} &55.54 &59.95 &64.62 &67.56 &44.82 &47.67 &47.73 &48.00\\ % Content row 4
MCD~\cite{MCD} &47.53 &52.60 &57.68 &59.95 &36.29 &40.08 &41.13 &42.05 \\ % Content row 5
\midrule % In-table horizontal line
TA$^{2}$N~\cite{TAN}  &57.14 &60.08 &64.09 &65.02 &42.39 &44.82 &44.39 &45.66\\
TA$^{3}$N~\cite{TAN}  &56.61 &62.35 &63.02 &63.95 &43.25 &44.27 &44.02 &42.05\\
\midrule % In-table horizontal line
\textbf{Semi-ABG} &57.28 &64.35 &65.42 &68.36 &59.80 &62.19 &62.46 &63.67\\
\textbf{Semi-HABG} &\textbf{61.15} &\textbf{65.29} &\textbf{67.29}&\textbf{70.36} &\textbf{60.39} &\textbf{62.54} &\textbf{63.36} &\textbf{63.98}\\ % Summary/total row
\midrule
Target Only &54.21 &58.88 &61.55 &66.62 &39.96 &42.54 &44.58 &44.36\\
\bottomrule \vspace{-1cm}
\end{tabular}
}
 % Table caption, can be commented out if no caption is required
\label{tab:Kinetics-Gameplay} % A label for referencing this table elsewhere, references are used in text as \ref{label}
\end{table}

\subsection{Semi-supervised Learning}
To study the robustness of the proposed algorithm, we extend the unsupervised domain adaptation to a semi-supervised setting, where a part of target labels are available for training. Extensive experiments are conducted on the most challenging ``Synthetic-to-Real'' testbed, \textit{i.e.}, the \textbf{Kinetics-Gameplay} dataset, on which the results with varying ratios of target labels are reported in Table \ref{tab:Kinetics-Gameplay}. The seen ratio of target labels ranges from $0.3$ to $0.9$. Similarly, \textbf{Source Only} / \textbf{Target only} represents the backbone model trained with source / target data only. All image-level domain adaptation methods (line 2-5), basic classification model (line 1,10) and our models (line 8-9) utilize \textbf{AvgPool} as the frame aggregation function. \textbf{TA$^2$N} and \textbf{TA$^3$N} use the \textbf{TRN} aggregator, since TRN is the major part of their works. It can bee seen that the overall performance is lower on the \textbf{Gameplay$\rightarrow$Kinetics} compared to \textbf{Kinetics$\rightarrow$Gameplay}, due to the insufficient samples in the source domain. The proposed \textbf{Semi-ABG} and its hierarchical variant \textbf{Semi-HABG} achieve higher recognition accuracy on both two transfer tasks, since the integrated graphs help to propagate the label information and take a full advantage of the supervision. The respective recognition accuracies of the proposed HABG on two transfer takss are improved by up to $8.0\%$ and $39.6\%$ over the state-of-the-art \textbf{TA$^3$N} with only $30\%$ of target labels available.

\vspace{-0.3cm}
\subsection{Ablation Study}
To investigate the validity of the derived modules and objective functions, we compare the four variants of \textbf{ABG} model on the full UCF-HMDB dataset. The comparison results are summarized in Table \ref{tab:ablation}. Removing the bipartite graphs, the \textbf{ABG w/o Graph} suffers a drop dramatically compared with the full model. The \textbf{ABG w/o $\mathcal{L}_d$} is the variant without the conditional adversarial learning, decreasing the recognition accuracy by $8.4\%$ and $2.95\%$ on average for the \textbf{UCF$\rightarrow$HMDB} and\textbf{ HMDB$\rightarrow$UCF} tasks, respectively. The \textbf{ABG w/o $\mathcal{L}_y^t$} refers to the variant without the entropy loss for target data, which triggers a slight decrease on the model performance. \textbf{HABG} is the hierarchical variant of the plain ABG model, performing better on the challenging UCF$\rightarrow$HMDB transfer task, which is consistent with the findings in semi-supervised experiments. The \textbf{ABG} model is more versatile and suitable for small datasets and easier transfer tasks such as HMDB$\rightarrow$UCF.

\begin{table}[t]
\caption{The ablation performance of the proposed ABG and HABG models on the full UCF-HMDB dataset.}\vspace{-0.4cm}
\resizebox{1\linewidth}{!}{% 
\begin{tabular}{l c c c c} 
\toprule % Top horizontal line
& \multicolumn{2}{c}{\textbf{UCF$\rightarrow$HMDB}$_{full}$} & \multicolumn{2}{c}{\textbf{HMDB$\rightarrow$UCF}$_{full}$}\\ 
\cmidrule(l){2-3}\cmidrule(l){4-5}
% Horizontal line spanning less than the full width of the table - you can add (r) or (l) just before the opening curly bracket to shorten the rule on the left or right side
\textbf{Method} &AvgPool &TRN &AvgPool &TRN\\ % Column names row
\midrule % In-table horizontal line
ABG w/o Graph &71.39 &73.89 &74.96  &74.61\\
ABG w/o $\mathcal{L}_d$ &72.78 &70.00 &82.14 &79.86\\
ABG w/o $\mathcal{L}_y^t$  &78.33 &75.83 &82.67 &81.79\\
HABG &\textbf{80.00} &\textbf{76.94} &82.49 &80.21\\
ABG &79.17 &76.67 &\textbf{85.11} &\textbf{81.79}\\
\bottomrule \vspace{-0.5cm}
\end{tabular}
}
\label{tab:ablation}
\end{table}

\vspace{-0.3cm}
\subsection{Parameter Sensitivity}
\begin{figure}[t]
    \centering
    \includegraphics[width=1\linewidth]{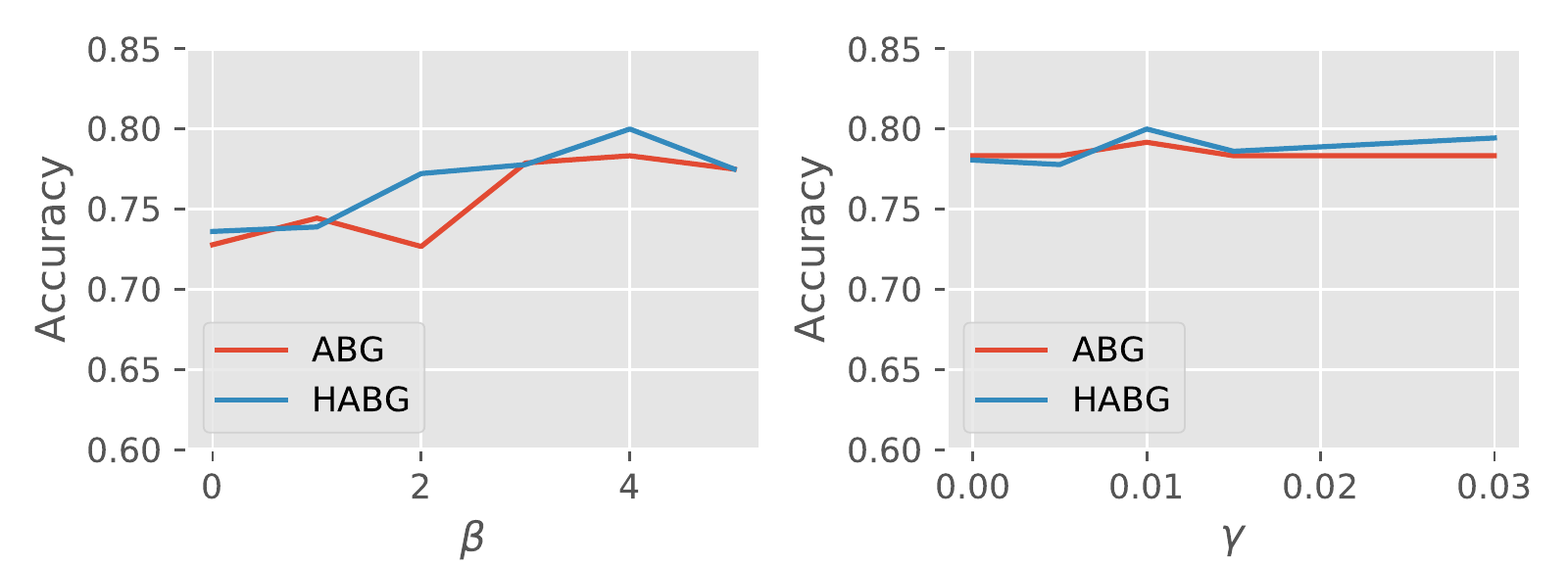}\\
    \includegraphics[width=1\linewidth]{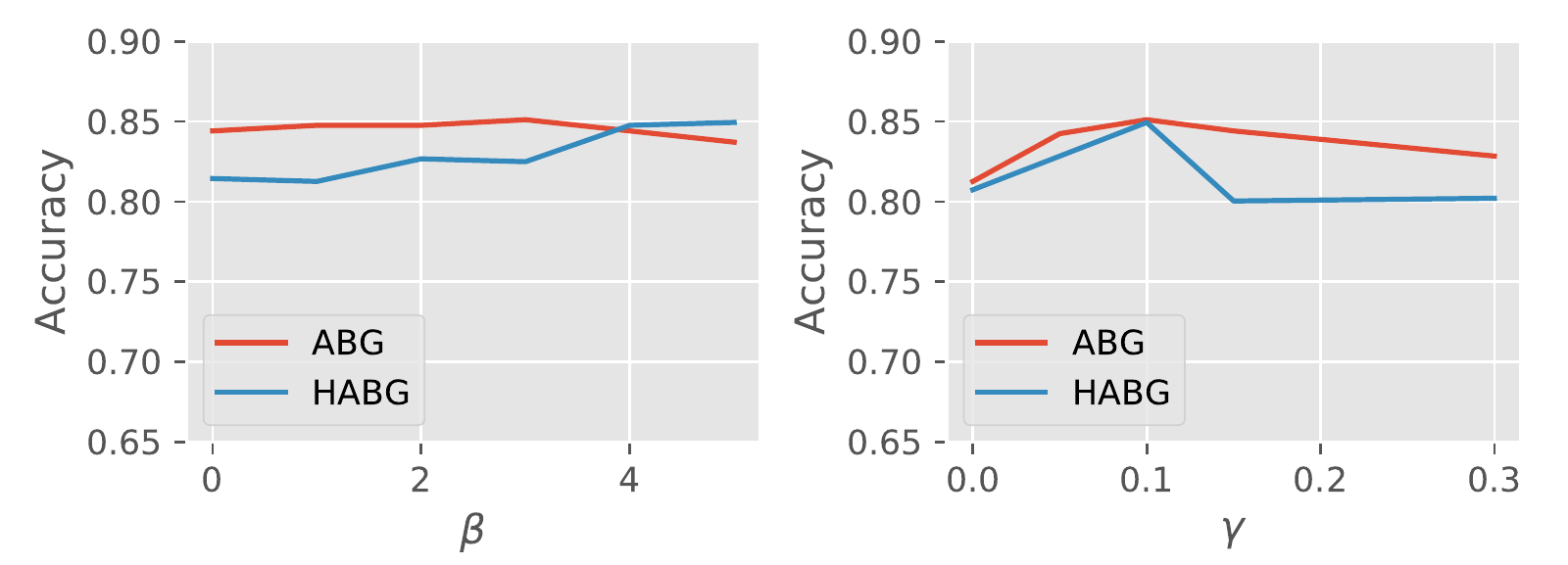}
    \caption{Performance comparisons of the proposed ABG and HABG with respect to the varying loss coefficients on the UCF$\rightarrow$HMDB (shown in the upper row) and HMDB$\rightarrow$UCF (shown in the bottom row) tasks.}
    \label{fig:params}
\end{figure}

To study the effect of the loss coefficients, we conduct the experiments on the UCF-HMDB$_{full}$ dataset with the varying values of $\beta$ and $\gamma$. The $\beta$ and $\gamma$ are utilized to reconcile the adversarial loss and the entropy loss, respectively. We compare the proposed ABG model and its variant HABG integrated with the identical AvgPool frame aggregation function. As plotted in Figure \ref{fig:params}, the average accuracies for both \textbf{ABG} and \textbf{HABG} models become quite stable when reaching sufficiently large loss coefficients. This indicates that our framework is robust with respect to loss coefficients.

\vspace{-0.2cm}
\subsection{Visualization}
To shed a qualitative light on evaluating the proposed model with various aggregation functions, we conduct the experiments on the HMDB$\rightarrow$UCF task, and visualize the features with t-SNE in Figure \ref{fig:tsne}. The features are extracted from the last layer of \textbf{ABG} and the baseline models, including \textbf{DANN}, \textbf{JAN}, \textbf{MCD} and \textbf{TA$^3$N}. Different colors indicate different classes. Circles represent the source videos and triangles represent the target videos. It is clearly shown that the features from ABG achieve the tighter clusters compared to the baselines.%, which is due to positive contributions of the bipartite graph and conditional adversarial module, which fuse the source and target frames locally and aligns the class-conditional distributions globally.

